\documentclass[sigconf]{acmart}

\author{Xuanqi Gao}
\affiliation{
  \institution{Xi'an Jiaotong University}
  \city{Xi'an}
  \country{China}
}
\email{gxq2000@stu.xjtu.edu.cn}

\author{Juan Zhai}
\affiliation{
  \institution{Rutgers University}
  \country{United States}
}
\email{juan.zhai@rutgers.edu}

\author{Shiqing Ma}
\affiliation{
  \institution{Rutgers University}
  \country{United States}
}
\email{shiqing.ma@rutgers.com}

\author{Chao Shen}
\affiliation{
  \institution{Xi'an Jiaotong University}
  \city{Xi'an}
  \country{China}
}
\email{chaoshen@mail.xjtu.edu.cn}

\author{Yufei Chen}
\affiliation{
  \institution{Xi'an Jiaotong University}
  \city{Xi'an}
  \country{China}
}
\email{yfchen@sei.xjtu.edu.cn}

\author{Qian Wang}
\affiliation{
  \institution{Wuhan University}
  \city{Wuhan}
  \country{China}
}
\email{qianwang@whu.edu.cn}

\usepackage{tikz}
\usepackage{amsmath}
\usepackage{filecontents}
\usepackage[flushleft]{threeparttable}
\usepackage{amsmath,amsfonts}
\usepackage{graphicx}
\usepackage{textcomp}

\usepackage[normalem]{ulem}

 \usepackage{url}

\usepackage{lipsum,tabularx}

\usepackage{multicol}
\usepackage{multirow}

\usepackage{colortbl}
\usepackage{booktabs}
\usepackage{setspace}

\hypersetup{
  linkcolor={blue!70!black},
  citecolor={red!70!black},
  urlcolor={blue!70!black}
}
\def\Snospace~{\S{}}

\usepackage[T1]{fontenc}

\usepackage{algorithm}
\usepackage{algorithmicx}
\usepackage[noend]{algpseudocode}

\usepackage{balance}

\usepackage{bm}
\usepackage{fp}
\usepackage{siunitx}
\usepackage{amsthm}

\usepackage[labelfont=bf,font=small,skip=5pt]{caption}
\usepackage{subcaption}

\usepackage{comment}

\usepackage{fancyhdr}
\usepackage{framed}
\colorlet{shadecolor}{blue!20}

\usepackage{tikz}

\usepackage{xspace}

\newcommand{\boxbeg}{
  \vspace{2px}
  \noindent\begin{tabular}{|l|}\hline
    \begin{minipage}{3.2in}
      \vspace{2px}
      \noindent
      }

      \newcommand{\boxend}{
      \vspace{2px}
    \end{minipage} \\ \hline
  \end{tabular}
  \vspace{-10pt}
}

\AtBeginDocument{%
  \providecommand\BibTeX{{%
    \normalfont B\kern-0.5em{\scshape i\kern-0.25em b}\kern-0.8em\TeX}}}
    
\hypersetup{draft}
\begin{document}

\copyrightyear{2022}
\acmYear{2022}
\setcopyright{acmlicensed}\acmConference[ICSE '22]{44th International
Conference on Software Engineering}{May 21--29, 2022}{Pittsburgh, PA, USA}
\acmBooktitle{44th International Conference on Software Engineering (ICSE
'22), May 21--29, 2022, Pittsburgh, PA, USA}
\acmPrice{15.00}
\acmDOI{10.1145/3510003.3510087}
\acmISBN{978-1-4503-9221-1/22/05}

\newcommand{\sys}{\mbox{\textsc{FairNeuron}}\xspace}
\title{\sys: Improving Deep Neural Network Fairness with Adversary Games on Selective Neurons}

\begin{abstract}
With Deep Neural Network~(DNN) being integrated into a growing number of critical systems with far-reaching impacts on society, there are increasing concerns on their ethical performance, such as fairness.
Unfortunately, model fairness and accuracy in many cases are contradictory goals to optimize during model training.
To solve this issue, there has been a number of works trying to improve model fairness by formalizing an adversarial game in the model level.
This approach introduces an adversary that evaluates the fairness of a model besides its prediction accuracy on the main task, and performs joint-optimization to achieve a balanced result.
In this paper, we noticed that when performing backward propagation based training, such contradictory phenomenon are also observable on individual neuron level.
Based on this observation, we propose \sys, a DNN model automatic repairing tool, to mitigate fairness concerns and balance the accuracy-fairness trade-off without introducing another model.
It works on detecting neurons with contradictory optimization directions from accuracy and fairness training goals, and achieving a trade-off by selective dropout.
Comparing with state-of-the-art methods, our approach is lightweight, scaling to large models and more efficient.
Our evaluation on three datasets shows that \sys can effectively improve all models' fairness while maintaining a stable utility.
\end{abstract}


\maketitle

\section{Introduction}\label{sec:intro}

Deep neural networks~(DNNs) are gradually adopted in a wide range of applications, including image recognition~\cite{heDeepResidualLearning2016}, self-driving~\cite{bojarskiEndEndLearning2016}, and natural language processing~\cite{huConvolutionalNeuralNetwork2014, kalchbrennerConvolutionalNeuralNetwork2014}.
One of the most trendy applications is decision-making systems, which requires a high utility DNN with fairness. 
As examples, artificial intelligent~(AI) judge~\cite{NABTurnsAI2020} or human resource~(HR)~\cite{HowAIWill2021} try to judge who should get a loan or interview. 
These systems should provide objective, supposedly consistent decision based on the given data, although there are often societal bias in these data~\cite{tramerFairTestDiscoveringUnwarranted2017}. 
We wish these systems could counteract unfair decision made by humans, but they still exhibit unfair behavior which affects individuals belonging to specific social subgroups. 
The COMPAS system is an example. 
It predicts recidivism of pretrial offenders~\cite{chouldechovaFairPredictionDisparate2017}, and continues to make decisions that favor Caucasians compared to African-Americans.
Such bias has made very negative societal impacts.
Therefore, it is crucial to have systematical methods for automatically fixing fairness problems in a given DNN model.

Intuitively, fairness problems happen when a model tends to make different decision for different instances which only differentiated by some sensitive attributes, such as age, race and gender.
Depending on specific tasks, the protected or sensitive attributes can vary.
Similarly, there are different fairness notations defined in existing DNN literature, e.g., group fairness~\cite{feldmanCertifyingRemovingDisparate2015}, individual fairness~\cite{dworkFairnessAwareness2012}, and max-min fairness~\cite{hashimotoFairnessDemographicsRepeated2018,lahotiFairnessDemographicsAdversarially2020}.
According to existing study~\cite{chouldechovaFairPredictionDisparate2017, kleinbergAlgorithmicFairness2018}, these fairness definitions are correlated with each other.
In practice, we usually consider a few representative ones, i.e., demographic parity, demographic parity rate and equal opportunity.
In this paper, we also consider these.

Existing DNN training frameworks, e.g., TensorFlow and PyTorch~\cite{abadiTensorFlowSystemLargeScale2016, paszkePyTorchImperativeStyle}, have provided no support for fairness problems detection and fixing.
Some other works try to fix other model problems~\cite{zhangAUTOTRAINERAutomaticDNN2021, maMODEAutomatedNeural2018, maLAMPDataProvenance2017}.
There are existing fairness fixing frameworks, such as FAD~\cite{adelOneNetworkAdversarialFairness2019} and Ethical Adversaries~\cite{delobelleEthicalAdversariesMitigating2021} that try to provide such functionality.
Based on the observations that optimizing accuracy and fairness can be contradictory goals in training, these frameworks introduce an adversary that monitors the fairness of the current training.
When fairness issues are detected, they solve it by various methods, e.g., data augmentation, that is, leveraging the adversary model to generate adversary examples which help fix the unfair problem and using them as part of the new training data.
Similar to generative adversary networks, training such an adversary can be time-consuming and challenging.
It has a lot of practical problems such as mode collapse~\cite{cheModeRegularizedGenerative2017}, which is hard to solve.
Moreover, such methods usually require using a more complex model training protocol, which is heavyweight.

We observe that the essential challenge of fixing model fairness is that optimization on accuracy only can lead to the selection of the usage of sensitive attributes.
For example, an AI HR that uses the sensitive attributes gender as an important feature will be biased.
Moreover, such feature selection happen in certain neurons or paths, which is different form the ones using all features or distinguishable features.
And such paths/neurons take a small portion of the whole network, otherwise, the network will have low accuracy for all samples.
Based on our observations, we proposed \sys, a fairness fixing algorithm that detects and repairs potential DNN fairness problems.
It works by first identify \textit{conflict paths} with a neural network slicing technique.
Conflict paths refer to the paths that contain a lot of neurons that select sensitive attributes to make predictions rather than distinguishable ones.
Then, we leverage such paths to cluster samples by measuring if they can trigger the selection of sensitive attributes.
Lastly, we retrain the model by selective retraining.
That is, for samples that can cause the model to select sensitive attributes as main features to make predictions, we enforce the DNN to reconsider this by muting other neurons that are not in the conflict paths.
By doing so, the conflict path neurons have to consider all features, otherwise, it will very low accuracy on other samples.
This helps remove the impacts of biased samples, and fix the fairness problem.

\sys has been implemented as a self-contained toolkit. 
Our experiments on three popular fairness datasets show that \sys improves twice fairness performance and takes one-fifth usage of training time on average than state-of-the-art solution, Ethical Adversaries~\cite{delobelleEthicalAdversariesMitigating2021}.
Note that \sys only relies on lightweight procedures like path analysis and dropout, which makes it much more effective and scalable than existing methods.

In summary, we make the following main contributions:
\begin{itemize}
    \item We propose a novel model fairness fixing frameworks. It avoids training an adversary model, and does not require modifying model training protocol or architecture. 
    It also features lightweight analysis and fixing, leading to high efficiency repairing.
    \item We develop a prototype \sys based on the proposed idea, and evaluate it with 3 popular public datasets. 
    The evaluation results demonstrate that \sys can effectively and efficiently improve fairness performance of models while maintaining a stable utility. 
    On average, the fairness performance DP can be improved by 57.65\%, which is 20\% higher than that of state-of-the-art adversary training based method, Ethical Adversaries.
    \item Our implementation, configurations and collected datasets are available at~\cite{gaoFairNeuron2022}.
\end{itemize}

\smallskip\noindent
\textbf{Roadmap}.
This paper is organized as follows. Section~\autoref{sec:background} presents the necessary background on fairness notions and fixing algorithms. 
In Section~\autoref{sec:design}, we discuss \sys in detail. 
Section~\autoref{sec:eval} shows our experiment setup and results. 
We review related works in \autoref{sec:rw} and conclude this paper in Section~\autoref{sec:conclusion}. 

\smallskip\noindent
\textbf{Threat to Validity}. 
\sys is currently evaluated on 3 datasets, which may be limited.
Similarly, there are configurable parameters used in \sys, and even though our experiments show that they are good enough to achieve high fixing results, this may not hold when the size of model is significantly larger or smaller.
Besides, we assume that most samples activate a limited number of paths, and most paths are activated by samples with certain features.
This has been observed by existing works~\cite{wangInterpretNeuralNetworks2018, qiuAdversarialDefenseNetwork2019}.
We also empirically validate this assumption in \autoref{sec:effect}. 
However, it is possible that this assumption may not hold for some models.
To mitigate these threats, all the original and repaired training scripts, model architecture and training configuration details, implementation including dependencies, and evaluation data are publicly available at~\cite{gaoFairNeuron2022} for reproduction.

\section{Background and Motivation}\label{sec:background}


\subsection{Fairness}\label{sec:notions}

Depending on concrete task specifications, fairness can have different notations~\cite{catonFairnessMachineLearning2020}.
These notions can be categorized into two groups: individual fairness~\cite{dworkFairnessAwareness2012,zemelLearningFairRepresentations}, which measures if individuals in the dataset is treated equally by the learned model; and group fairness~\cite{feldmanCertifyingRemovingDisparate2015, hardtEqualityOpportunitySupervised2016}, which concerns about whether subpopulation with different sensitive attributes are treated equally.
For example, for an online shopping recommendation system, all customers in the dataset should be treated equally, which asks for individual fairness.
For an AI powered hiring system, applicants with sensitive attributes (e.g., different genders) should be treated equally, which is a typical case of group fairness.

Before discussing different fairness notations, we first define a set of notations. 
We denote the sensitive attribute as \(S\) and other observable insensitive attributes as \(A\). 
We assume that the subpopulation with \(S=1\) is the disadvantaged group, and the privileged group is the subpopulation with \(S=0\).
Also, we represent the true label as \(Y\), and the predicted output, i.e., positive/negative as \(\hat{Y}\) which is a random variable depending on attributes \(S\) and \(A\).
\(\hat{Y}=1\) and \(\hat{Y}=0\) are the positive and negative outcomes, respectively.
Following such notations, we can define commonly used different fairness notations as follows:

\smallskip
\noindent
\textbf{Demographic parity (DP)}. 
Demographic parity, or statistical parity, is one of the earliest definitions of fairness~\cite{dworkFairnessAwareness2012}. 
It views fairness as different subpopulations (i.e., \(S=0\) and \(S=1\)) should have an equal probability of being classified to the positive label.
Formally, demographic parity measures the probability differences between different groups: 
\begin{equation}
    DP = \left|\;P(\hat{Y}=1\;|\;S=0)-P(\hat{Y}=1\;|\;S=1)\;\right|
    \label{DP_notion}
\end{equation}

In an ideal case, we say that a model is when \(DP = 0\), which indicates that  the prediction output \(\hat{Y}\) and sensitive attribute \(S\) are statistically independent.
If so, the output is not affected by the sensitive attribute, and hence the model is not biased towards certain values of the sensitive attribute showing fairness in prediction.
In practice, \(DP=0\) is hard to get and we view a model as fair when \(DP \leq \epsilon\) where \(\epsilon\) is a threshold value that is determined by real world tasks and requirements.

\smallskip
\noindent
\textbf{Demographic parity ratio (DPR)}.
Demographic parity ratio, or disparate impact, is similar to demographic parity.
The key difference is that it represents the equality or similarly of prediction on different groups as a ratio (instead of a substitution).
Formally, it is defined as:
\begin{equation}
    DPR=\frac{P(\hat{Y}=1\;|\;S=1)}{P(\hat{Y}=1\;|\;S=0)}
    \label{DPR_notion}
\end{equation}

Like demographic parity, \(DPR = 1\) indicates a fair model in the ideal case, and in practice, we say a model is fair when \(DPR  \geq \tau \) where \( \tau \) is the fairness threshold.
Moreover, it also focuses on the probability of different groups being classified to the positive label. 
The key difference is that DPR measures the differences in a ratio.
This is because its origins are in legal fairness considerations for selection procedures which the Pareto principle, a.k.a., the 80\% rule, is commonly used~\cite{feldmanCertifyingRemovingDisparate2015}.
To make a direct comparison with 80\%, DPR calculated the ratio instead of substitution.

\smallskip
\noindent
\textbf{Equal opportunity (EO)}.
A limitation of DP and DPR is that they do not consider potential differences in compared subgroups. 
Equal opportunity (EO) overcomes this by making use of the FPR (false positive rate) and TPR (true positive rate) between subgroups~\cite{hardtEqualityOpportunitySupervised2016}.
Formally, EO is defined as:
\begin{equation}
    EO = \left|\;P(\hat{Y}=1\;|\;S=0,Y=1)-P(\hat{Y}=1\;|\;S=1,Y=1)\;\right|
    \label{EO_notion}
\end{equation}

A model achieves EO fairness when \(EO = 0\), namely, the prediction is (conditional) independent of the sensitive attribute \(S\).
In practice, we say an model is EO fair when \(EO \leq \nu\) and here, \(\nu\) is the fairness threshold value.

Besides these discussed notions, there are many other fairness definitions, such as fairness through unawareness (FTU)~\cite{kusnerCounterfactualFairness}, disparate treatment~\cite{barocasBigDataDisparate2016}, disparate mistreatment~\cite{zafarFairnessDisparateTreatment2017}, counterfactual fairness~\cite{kusnerCounterfactualFairness}, ex-ante fairness and ex-post fairness~\cite{freemanBestBothWorlds2020}, etc.
Friedler et al.~\cite{friedlerComparativeStudyFairnessenhancing2019a} compared different notations and measured their correlations on the Ricci and Adult datasets.
Results show that different notations have strong correlations with each other.
As a result, most work usually pick a few representative ones.
Following existing related work, we choose three most common notations, i.e., DP, DPR~\cite{dworkFairnessAwareness2012}, and EO~\cite{hardtEqualityOpportunitySupervised2016} in our study.




\subsection{Improving DNN Fairness}\label{sec:existing}

Many machine learning algorithms including DNNs suffer from the bias problem.
Namely, the model can make a decision based on wrong attributes. 
For example, a biased hiring AI may make admissions based on applicants' gender information.
Such issues can be caused by the biased training data or the algorithm itself.
DNN has shown to be a biased algorithm, and potentially trained DNN models can make unfair predictions despite its high accuracy.
This can lead to severe problems especially when DNNs are becoming more and more popular including applications like AI judge~\cite{NABTurnsAI2020}, AI based authentication, AI HR~\cite{HowAIWill2021}, etc. For example, COMPAS, a popular system that predicts the risk of recidivism, claimed that ``black people re-offend more''~\cite{MachineBiasProPublica}; the first beauty contest robot, Beauty.AI~\cite{FirstInternationalBeauty}, ``found dark skin unattractive''~\cite{BeautyContestWas2016}; and the Microsoft chatbot Tay became a racist and sex-crazed neo-Nazi~\cite{MicrosoftNeoNaziSexbot}.
Biased AIs in such systems can lead to severe ethical concerns, potentially threatening our daily life and economy. 
As a response to this issue, existing work has proposed methods to improve DNN fairness by removing such bias.

\smallskip
\noindent \textbf{FAD}.
Adel et al.~\cite{adelOneNetworkAdversarialFairness2019} proposed a fair adversarial framework FAD, which leverages gradient reversal~\cite{ganinUnsupervisedDomainAdaptation}~(which acts as an identity function during forward propagation and multiplies its input by -1 during back propagation) to fix model fairness problems.
The authors introduced an adversarial network to encode fairness into the model: a predictor network \(\mathcal{F}_P\) and an adversary network \(\mathcal{F}_A\).
The goal of the predictor is to maximize accuracy in \(\hat{Y}\) while the adversary network tries to maximize fairness in protected attribute \(S\).
For fairness fixing, we need a new model architecture which can: (i) predict the true label \(Y\), and (ii) not be able to predict the sensitive attribute \(S\):
\begin{equation}
    L_{\mathcal{F}_P} = L_{CE} - \alpha L_{\mathcal{F}_A}
    \label{joint_loss}
\end{equation}
where \(L_{\mathcal{F}_P}\), \(L_{CE}\),  \(L_{\mathcal{F}_A}\) denote the predictor loss, predictor logistic loss (a.k.a., CE loss) and the adversary logistic loss, respectively.
The hyperparameter \(\alpha\) regulates the accuracy-fairness trade off.
After that, it uses a post-training process to align TP (true positive) and FP (false positive) across all classes by adjusting class-specific threshold values of logits with a ROC analysis introduced by Hardt et al.~\cite{hardtEqualityOpportunitySupervised2016}.

\smallskip
\noindent \textbf{Ethical Adversaries}.
Delobelle et al.~\cite{delobelleEthicalAdversariesMitigating2021} proposed the ethical adversaries framework to solve the fairness problem.
The framework has two parties, the external adversary, a.k.a., the feeder, and the reader, which represents the protected attribute \(S\).
It is an iterative training procedures, during which each party interacts with each other.
The reader is trained with the target label at the same time, and each time, it evaluates if the training has bias or not.
If so, it propagates the related gradient back to the network.
The feeder can be viewed as a data augmenter which performs evasion attacks to find adversarial examples that can be used in the adversarial training.
During this adversarial training, the target label (i.e., main task of the model) and the fairness goal is adjusted by a hyperparameter \(\lambda\), which is similar to the FAD framework.
    
\smallskip
\noindent \textbf{Pre-/Post-processing Methods}.
FAD and Ethical Adversaries are online methods which solves the fairness issue during training.
There are other methods that leverages pre-processing or post-processing to solve this problem, e.g., reweighing ~\cite{kamiranDataPreprocessingTechniques2012}, and reject option classification~(ROC~\cite{kamiranDecisionTheoryDiscriminationAware2012}).
Reweighing assigns different weights to input samples in different group to make the dataset discrimination-free (pre-processing).
ROC gives favorable outcomes to unprivileged groups and unfavorable outcomes to privileged groups in a confidence band around the decision boundary with the highest uncertainty (post-processing).





\subsection{Motivation and Basic Idea}\label{sec:ours}



\noindent
\textbf{Limitations of existing work.}
Existing work has a few limitations.
Firstly, they introduce another model as the adversary in the training procedure. 
Inheriting from existing adversarial networks, training such models is not easy. 
Problems like mode collapse~\cite{cheModeRegularizedGenerative2017}, failing to converge~\cite{mertikopoulosCyclesAdversarialRegularized2018}, and vanishing gradients are quite common in such a model structure. 
This will require extra efforts in solving such problems.
Secondly, there is no guarantee that training such adversary networks will always converge for now.
There is a theoretical guarantee that GAN (generative adversary network) will converge, despite its practical difficulty.
As a minimax game, training such GAN models will converge when it achieves the Nash equilibrium~\cite{goodfellowGenerativeAdversarialNets2014}.
However, FAD and Ethical Adversaries empirically observe that model accuracy and fairness may conflict with each other in some cases and may not conflict with each other in other cases.
On one hand, this shows that there exist models that are both accurate and fair.
On the other hand, it also indicates that the designed adversary training is not a zero-sum game, and there is no guarantee to show the existence of Nash equilibrium in this game.
As a result, existing work can fail to converge when training the model because the game has no solutions.
Empirical results confirm this conclusion.
\autoref{sec:effect} reports that FAD may exacerbate fairness problem. 
As \autoref{tab:effective} shown, FAD results in decreasing of DPR on Census and increasing of EO on COMPAS, which mean the fairness problems has not been mitigated from these perspectives.
Elazar and Goldberg made an empirical observation on leakage of protected attributes for neural networks trained on text data, which can also demonstrate this conclusion~\cite{elazarAdversarialRemovalDemographic2018}.

\smallskip
\noindent
\textbf{Why bias happens in a DNN training?}
Based on existing literatures and our experiences, we make a few key observations that are important for us to develop our method.

\smallskip
\noindent
\textit{\uline{Observation I: Optimizing accuracy and different fairness objectives can be contradictory to each other, but not always.}}
Existing work~\cite{delobelleEthicalAdversariesMitigating2021,  adelOneNetworkAdversarialFairness2019} has shown that accuracy and different fairness goals (e.g., DP, DPR and EO), including different fairness goals themselves, can be contradictory to each other.
This is the reason why some models with high accuracy are highly biased: when optimizing during training, directions with higher accuracy gain may be contradictory to directions with higher fairness.
The good news is that existing work has empirically demonstrate that it is possible to train a model with high accuracy and fairness at the same time~\cite{delobelleEthicalAdversariesMitigating2021}.

\smallskip
\noindent
\textit{\uline{Observation II: A neuron represents a combination of different features, and model bias indicates that the model focuses on certain features that it should not.}}
As a general understanding of DNNs, each neuron in the network is extracting features from the input.
Form the input layer to the final prediction layer, the extracted features are becoming more and more abstract.
Each neuron is representing a set of features it receives from the previous layer, and weights can help it determine which set of features are more important compared with others.
A model is biased indicates that a model is focusing on the wrong features, e.g, AI judges should be affected by sensitive attributes like genders.
For example, a hiring AI is biased on gender when it selects gender feature rather than others as an important factor to decide if a candidate can get an interview.
Notice that such importance is represented by weights in the DNN.

Now, we can use our observations to explain why bias happens in training a deep neural network.
When training a DNN, the optimizer tries to pick important features based on gradient information.
When updating individual neurons, it may encounter cases where the fairness and accuracy optimization subjects are pointing to different directions.
If it only considers accuracy as its training goal, it will select the direction that optimizes the accuracy the most which can lead to low fairness.
Furthermore, we know that the gradient information is calculated based on given samples.
If we are able to detect such samples, we can potentially fix the problem by enforcing the optimizer to pick the correct set of features.

\smallskip
\noindent
\textbf{Our idea.}
Based on our observations, we argue that it is not necessary to introduce an adversary that detects the potential conflicts of optimizing the accuracy and fairness.
Instead, we first monitor the training process to detect neurons whose accuracy and fairness optimization get conflicts with each other.
Then, we identify samples that causes such contradictory optimizations.
Lastly, we enforce the optimizer to decide a balanced optimal direction that optimizes both accuracy and fairness.
By doing so, we remove the need of introducing an adversary.
It simplifies the training procedure and is more lightweight compared with existing solutions.

\section{Design}\label{sec:design}

\subsection{Overview}\label{sec:overview}

\begin{figure*}[h]
  \centering
  \includegraphics[trim={0 90 0 200},clip,width=0.95\linewidth]{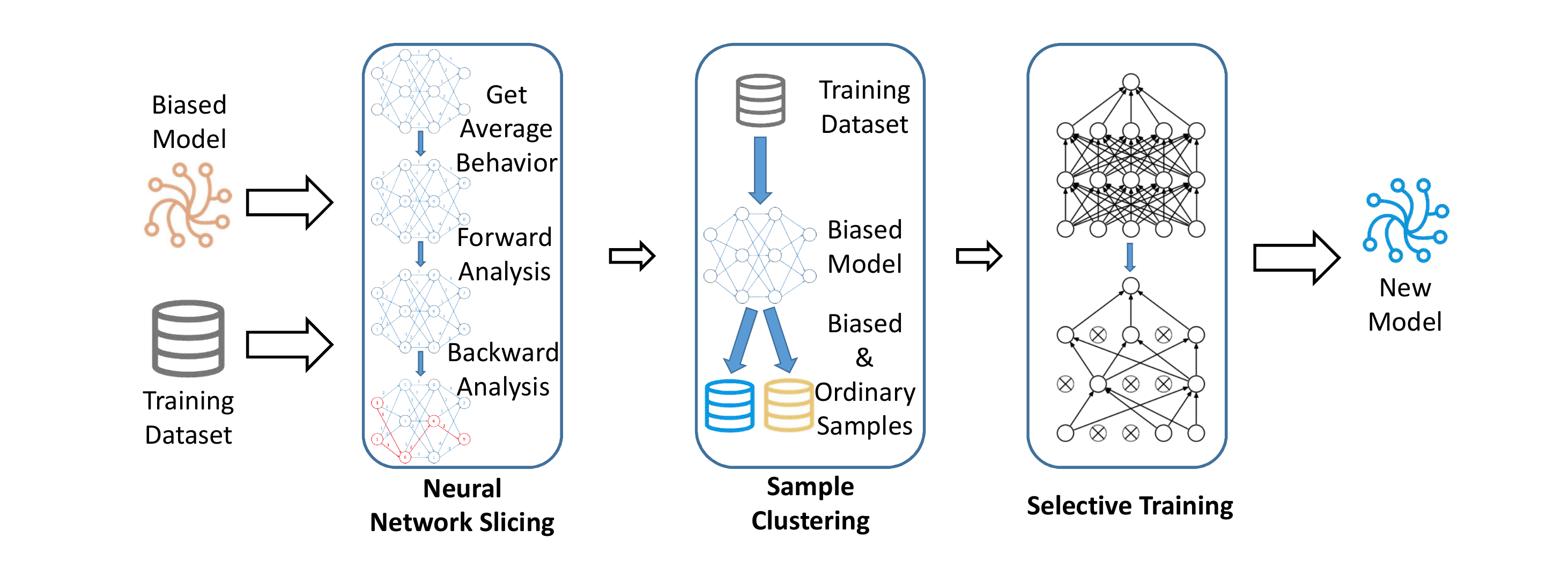}
  \caption{Overview of \sys.}\label{fig:overview}
  \Description{Overview of \sys.}
\end{figure*}

\textbf{Workflow.}
\autoref{fig:overview} presents the workflow of \sys.
It takes a biased model and its training data as inputs, and outputs a fixed model.
Firstly, \sys performs \textit{neural network slicing}, which detects neurons and paths that have contradictory optimization directions.
Notice that because of the dense connections of DNN, such neurons are typically connected with each, passing the biased features from one layer to the next.
As such, we do this in the path granularity.
In this step, we leverage a neuron slicing technique which performs a differential analysis to identify the target paths.
Next, we leverage such paths to identify the samples that cause such effects, known as the \textit{sample clustering}.
After this step, we can separate the samples into two clusters, biased data samples and benign samples.
Lastly, we perform \textit{selective retraining} to enforce the model to learn unbiased features.
Essentially, for samples in different clusters, we have different training strategies.
For samples in the benign data cluster, we do not change anything, while for samples in the biased cluster, we enforce detected neurons to consider a larger set of features and weigh them to learn {\it all} features that are important for prediction rather than the biased ones.




\smallskip\noindent
\textbf{Algorithm.}
The overall algorithm of \sys is presented in \autoref{alg:overview}, denoted as Procedure \texttt{\sys}.
As mentioned before, it takes a biased model \textit{BiasedModel} and training dataset \textit{TrainDataset} as inputs, and outputs a fixed model, referred to as \textit{NewModel} in the algorithm.
In the main algorithm, \sys analyze the relationship between dataset and model by getting activation paths of each input sample~(line 1-5). 
After acquiring path information, \sys groups the training dataset into two parts~(line 6). 
The first one is consists of benign samples whose activation paths are clustered by samples, and the second one is consists of biased samples and corresponding paths.
Then \sys performs different training strategies on them, it deactivates dropout layers when training benign samples and activated them when training biased samples.

\begin{algorithm}
  \caption{\sys Algorithm}\label{alg:overview}
  \begin{algorithmic}[1]
    \Require \(BiasedModel\): a biased model to fix 
    \Require \(TrainDataset\): training dataset
    \Ensure \(NewModel\): trained model after fixing
    \Procedure{\sys}{}
    \State $PathList \gets []$
    \For{$sample \in TrainDataset$}
    \State $P \gets GetActivationPath(BiasedModel,sample,\gamma)$
    \State $P.sample \gets sample$
    \State $Append(PathList,P)$
    \EndFor
    \State $O,S \gets GetSamplesDivided(PathList,\theta)$
    \State $NewModel \gets BiasedModel$
    \For{$o \in O$}
    \State $OrdinaryTraining(NewModel,o)$
    \EndFor
    \For{$s \in S$}
    \State $DropoutTraining(NewModel,s)$
    \EndFor
    \Return $NewModel$
    \EndProcedure

    \item[]
    \Require \(Model\): model to analyze
    \Require \(Sample\): samples used in analyze
    \Require \(\gamma\): hyperparameter to determine the activation of neurons
    \Ensure $P$: path activated
    \Procedure{GetActivationPath}{}
    \State $P \gets \emptyset$
    \State $Q \gets \emptyset$
    \State $Q \gets OutputNeuron$
    \While{$Q \neq \emptyset$}
      \State $Q' \gets \emptyset$
      \For{$q \in Q$}
        \State $N \gets GetPreNeuron(q)$
        \For{$n \in N$}
          \State $ContribList[n] \gets ComputeContrib(n)$
        \EndFor
        \State $SortedList\gets Sort(ContribList)$
        \State $Sum \gets 0$
        \For{$i \gets 0 \  to \  len(ContribList)$}
          \If{$Sum \leq \gamma \times q.value$}
            \State $Sum \gets Sum + SortedList[i].value$
            \State $Q' \gets Q' \cap SortedList[i].index$
            \State $P \gets P \cap (SortedList[i].index,q.index)$
          \EndIf
        \EndFor
      \EndFor
      \State $Q \gets Q'$
    \EndWhile
    \Return $P$
    \EndProcedure

    \item[]
    \Require \(PathList\): a list of paths used to cluster samples
    \Require \(\theta\): hyperparameters used to find conflict paths
    \Ensure \(OL, AL\): benign and biased samples, respectively.
    \Procedure{GetSamplesDivided}{}
    \State $OL \gets []$
    \State $AL \gets []$
    \State $PathList.count \gets Count(PathList.samples)$
    \State $M \gets Max(PathList.count)$
    \For{$i \gets 0 \ to \ len(PathList)$}
      \If{$PathList[i].count \leq \theta \times M$}
        \State $Append(OL,PathList[i].samples)$
      \Else
        \State $Append(AL,PathList[i].samples)$
      \EndIf
    \EndFor
    \Return $OrdinaryList,AbnormalList$
    \EndProcedure
  \end{algorithmic}
\end{algorithm}

\subsection{Neural Network Slicing}\label{sec:path_analysis}

\begin{figure*}[h]
  \centering
  \includegraphics[trim={0 70 0 50},clip,width=0.95\linewidth]{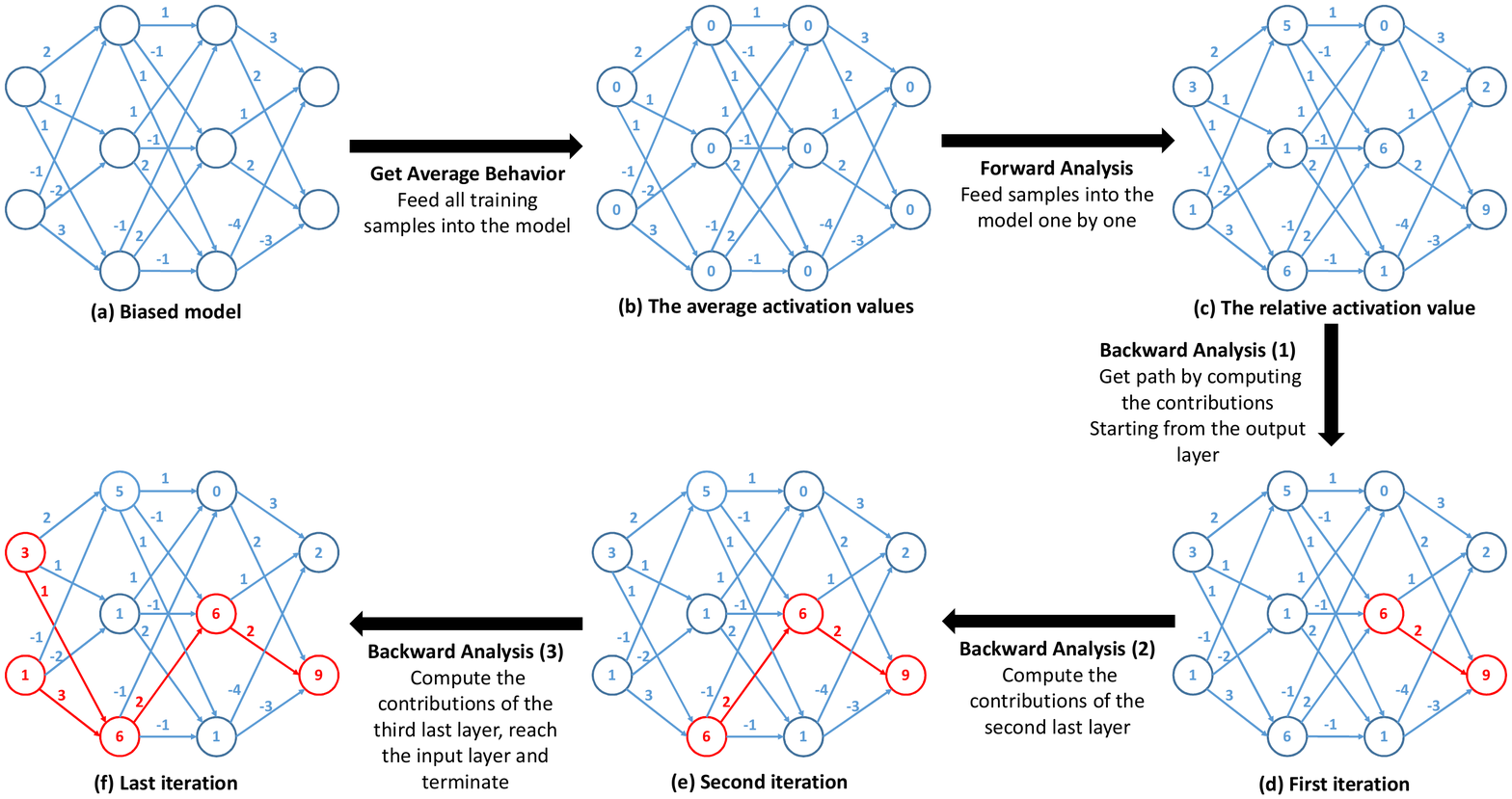}
  \caption{Abnormal path detection example.}
  \Description{Abnormal path detection example.}
  \label{fig:path_analysis}
\end{figure*}

In Neural Network Slicing, we try to find paths and neurons that contain the optimizer finds contradictory optimization directions for accuracy and fairness.
\autoref{fig:path_analysis} shows the neural network slicing method of \sys. 
The input of this algorithm is the training dataset and the biased model to fix, a neural network which has already learned the weights based on a training dataset. 
We will use this example to demonstrate how it works in this section.

First, \sys gets the \textit{average behavior of a neuron}, by leveraging its activation values.
The behavior of a neuron can be represented in many ways, and in \sys, we use the most simple and naive representation, its averaged activation value.
Specifically, \sys calculates the average activation values of the neuron for a given dataset, which is usually the training dataset.

Then, \sys performs a \textit{forward analysis} to understand the diversity of a neuron behavior.
Similar to the first step, we also use the activation values of a neuron to represent the behavior of the neuron.
In this step, we feed individual inputs to the DNN, and record the activation value differences between the average activation and the value for this concrete input.
By doing so, we can estimate the contributions of each neuron to the output for a given sample.
This helps us to identify neurons that contain biased features.

Afterwards, we obtain paths that contain biased features.
Notice that a DNN is a highly connected network, and as a layered structure, behaviors of a single layer will be passed to the next layer.
Because of this, biased features will be accumulated in this network, and as a result, neurons in the last few layers will contribute a lot to the biased prediction.
On the other hand, these neurons do not denote the root cause of such bias.
To completely fix the neuron network, it is important to identify the whole chain of such propagation.
So we comprehensively consider neurons and synapses and calculate their contributions, and backtrack these contributions in the network.
We show the detail of this phase in the procedure \textit{GetActivationPath} in \autoref{alg:overview}.
Starting from the output neuron, we iteratively compute the contributions of the previous neurons~(line 19-22), which is similar to the backward propagation.
Then, we sort them in descending order~(line 23), and add the key synapses and corresponding neurons into the path set~(line 24-29).
To determine if a synapse is a key synapse or not, we need to calculate whether the sum of all the synapses that are connected to the same successor neuron is still less than the threshold. 
The threshold is determined by the activation value of subsequent neurons and the hyperparameter \(\gamma\).

Lastly, we identify conflict paths, namely paths that contain features causing the biased prediction.
Based on our observations, we know that when making predictions, the model uses the benign feature set to make predictions for benign samples and use the biased feature set to make predictions for biased samples.
Considering that a neuron represents a set of features, we know that biased samples are activating neurons/paths that are different from the others.
Notice that biased paths/neurons takes a relatively small portion of the whole neural network.
Otherwise, the network will make predictions on a lot of biased features, leading to low accuracy.
Based on this intuition, we obtain such conflict paths by analyzing the frequency of the activated paths.
More specifically, we set the activation frequency of the most frequently activated path as the standard, and compare the activation frequency of each path with it.
If the activation frequency of a certain path is less than a certain percentage of the standard~(determined by \(\theta\)), then it can be considered as a conflict path.

\smallskip\noindent
\textit{\uline{Example}}. Assume the biased model is a simple neural network shown in \autoref{fig:path_analysis}. 
The weight values have been labeled on the corresponding synapses in \autoref{fig:path_analysis}(a). First we perform path profiling, and the results are set to 0 for simplicity. 
Then we feed a sample (3,1) into the model, and calculate its relative activation value on each neuron. 
Take the top neuron of the second layer as an example, its relative activation value is $3\times 2 + 1 \times (-1) -0=5$, as shown in \autoref{fig:path_analysis}(c). Finally we backtrack the contributions of synapses to get the activation path. 
\autoref{fig:path_analysis}(d)-(f) shows how we get a path iteratively. 
Let us denote the $k$-th neuron in the $m$-th layer as $n^m_k$.
At first $Q={n^4_2}$, assuming $\gamma = 0.8$, we add $n^3_2$ into $Q'$ and do not add the others because $|6 \times 2| > |0.8 \times 9|$. 
Then we let $Q=Q'={n^3_2}$, and add $n^2_3$ since $|6\times 2| > |0.8 \times 6|$. Last we add $n^1_1$ and $n^1_2$ into $Q'$ because $|3 \times 1 \leq 0.8 \times 6|$ and $|3 \times 1|+|1 \times 3| > |0.8 \times 6|$. 
Ultimately we get all the paths iteratively. 
The conflict paths detection can be regarded as the preceding step of sample clustering, and the example in \autoref{sec:sorting} shows its process. \qed

\subsection{Sample Clustering}\label{sec:sorting}

The sample clustering aims to measure the impact of input samples on fairness.
After detecting conflict paths, we can distinguish these neurons exhibiting biased behavior.
We handle the corresponding samples of these neurons~(denoted as \textit{biased sample}) to improve their fairness performance. 
Since we recorded the relationship between paths and samples~(line 4 in \autoref{alg:overview}), we can easily find these corresponding samples and get the training dataset divided into two groups.

As shown in \autoref{alg:overview}, \(PathList\) is a list which contains each path and its corresponding activation samples. 
First, we count the total number of path's corresponding activation samples one by one, and the number is denoted as \textit{activation frequency}~(line 34). 
Second, we sort the activation frequency list we get above, and record its maximum as \(M\)~(line 35). 
Third, we check whether these paths' activation frequencies are greater than the threshold \(\theta \times M\).
We denote the paths which do not meet the above condition as \textit{biased path}, and denote their corresponding samples as \textit{biased sample}. 
After we detecting the biased paths, these biased samples can be separated from ordinary samples~(line 36-40).


\begin{figure}[]
  \centering
  \includegraphics[trim={0 0 0 0},clip,width=0.95\linewidth]{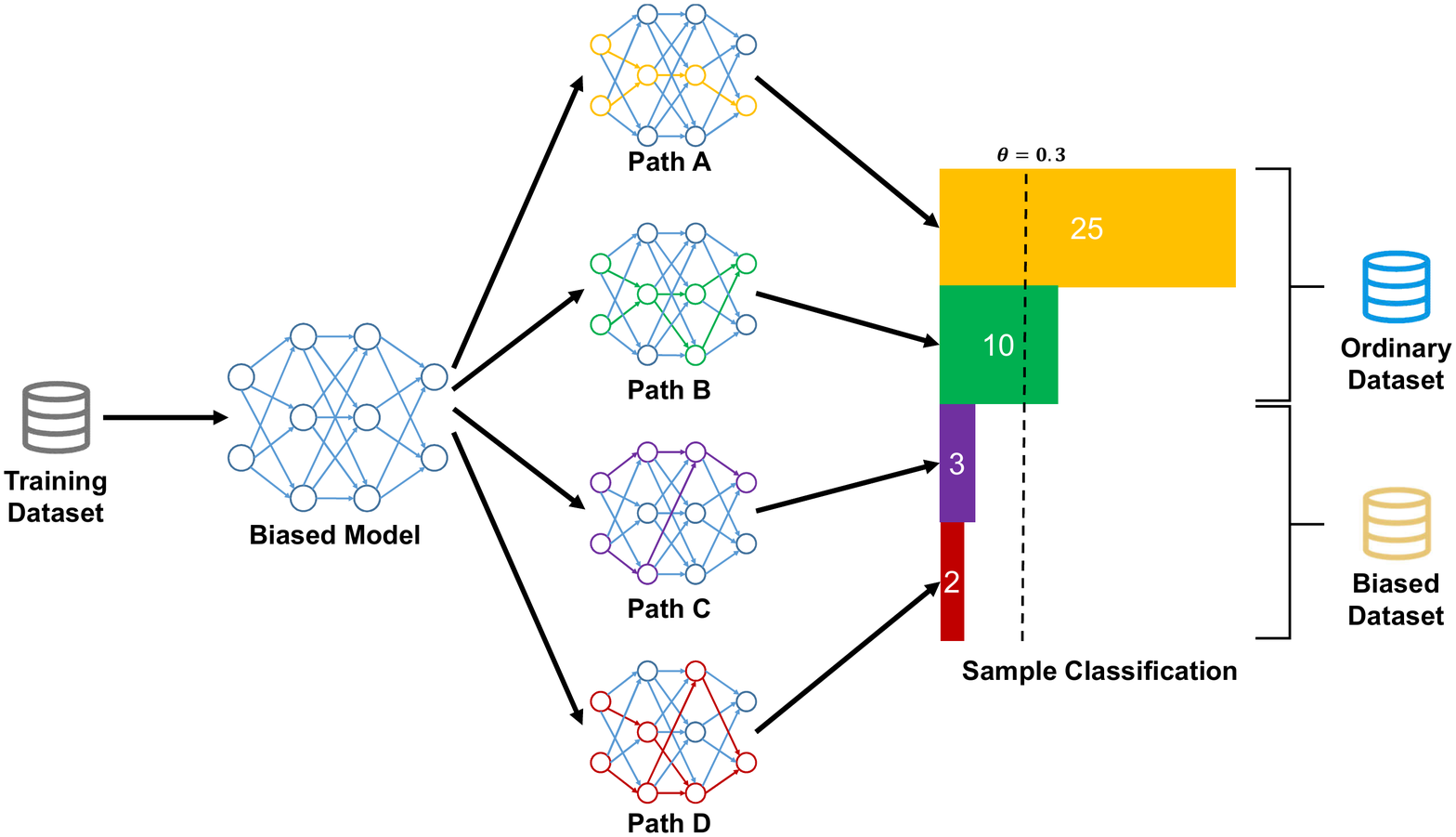}
  \caption{Sample clustering example.}
  \Description{Abnormal path detection example.}
  \label{fig:sample_clustering}
\end{figure}

\smallskip\noindent
\textit{\uline{Example}}.
Suppose our training dataset has 40 samples, as shown in \autoref{fig:sample_clustering}. 
We feed the training dataset into the biased model, and obtain 4 different paths A, B, C and D based on the procedure \textit{GetActivationPath} in \autoref{alg:overview}.
Then we count the number of samples activating these 4 paths, and get 25, 10, 3, and 2 for A, B, C, and D, respectively. 
We assume that \(\theta=0.3\), so the threshold is \(25\times 0.3=7.5\) since the maximum of path activation statistics is 25. 
Then, path C and D will both be classified as biased paths, which results in 3 samples activating path C and 2 samples activating path D being grouped in biased samples.
\qed

\subsection{Selective Training}\label{sec:training}

Finally, we perform ordinary training on the ordinary samples and dropout training on biased samples obtained above. 
We do not need to change the model structure, only need to change the activation state of the dropout layers.
Ordinary training means deactivating the dropout layers for training.
With the current training system, we can activate dropout layers when training on these biased samples and vice versa.
By performing dropout training on these biased neurons, we enforce them to learn more unbiased features rather than biased ones to mitigate the fairness problems.

\section{Evaluation}\label{sec:eval}

\subsection{Experiment Setup}\label{sec:setup}

\subsubsection{Hardware and software}\label{sec:envs}

We conducted our experiments on a GPU server with 32 cores Intel Xeon 2.10GHz CPU, 256 GB system memory and 1 NVIDIA TITAN V GPU running the Ubuntu 16.04 operating system.

\subsubsection{Datasets}\label{sec:datasets}

We evaluated our method on three popular datasets: the UCI Adult Census, COMPAS, and German Credit. 
\begin{itemize}
    \item \textbf{UCI Adult Census.} The UCI Adult Census was extracted from the 1994 Census bureau database, gathering 32,561 instances represented by 9 features such as age, education and occupation. The gender is considered as the sensitive attribute.
    \item \textbf{COMPAS.} The COMPAS system is a popular commercial algorithm used by judges for predicting the risk of recidivism, and the COMPAS dataset is a sample outcome from the COMPAS system.
    The race of each defendant is the sensitive attribute.
    \item \textbf{German Credit.} This is a small dataset with 600 records and 20 attributes.
    The original aim of the dataset is to give an assessment of individual’s credit based on personal and financial records. The gender is the sensitive attribute.
\end{itemize}

\subsubsection{Models}\label{sec:model}

In our experiment, we built a fully-connected neural network with three hidden layers for each dataset respectively.
For the COMPAS and the German Credit dataset, each hidden layer is composed of 32 neurons, while for the UCI Adult Census dataset, each hidden layer is composed of 128 neurons due to its larger encoded input. The details of the models used in the experiments are shown in \autoref{tab:models}.

\begin{table}[]
        \centering
        \caption{Experimented DNN models.}
        \label{tab:models}
        \begin{tabular}{@{}llr@{}}
        \toprule
        \multicolumn{1}{c}{Dataset} & \multicolumn{1}{c}{Model}                             & \multicolumn{1}{c}{Accuracy} \\ \midrule
        Census  & 3 Hidden-layer Fully-connected NN & 83.9\%   \\
        Credit  & 3 Hidden-layer Fully-connected NN & 73.4\%   \\
        COMPAS  & 3 Hidden-layer Fully-connected NN & 62.1\%  \\ \bottomrule
        \end{tabular}
\end{table}

We use the softmax activation function for Census and German Credit to achieve binary classification, and the linear activation function for COMPAS to get recidivism scores. 
We randomly separate the dataset into the training, validation, and test sets, by a ratio of 7:1:2, respectively.
The neural network is trained by the Adam optimizer with $\beta_1 = 0.9$, $\beta_2 = 0.9999$, and initial learning rate $l_r = 0.01$, which is scheduled by a factor of 0.1 when reaching a plateau.

\subsubsection{Hyperparameter tuning}\label{sec:tuning}
To obtain the suitable hyperparameters $\theta$ and $\gamma$, we run a parallel grid search for hyperparameters to optimize training loss function. 
We sample $\theta$ between the interval $[10^{-4},1]$ proportionally, and sample $\gamma$ between the interval $[0.5,1]$. 
Following the standard practice in machine learning, the grid search is performed on a small subset drawn from the training set in a certain proportion~(e.g., 10\%), and we utilize the \textit{ray tune tool} to perform it automatically~\cite{TuneScalableHyperparameter}.

\subsubsection{Metrics and baseline methods}\label{sec:baseline}
We compare our algorithm with other popular in-processing state-of the-art fixing algorithms, such as FAD~\cite{adelOneNetworkAdversarialFairness2019} and Ethical Adversaries~\cite{delobelleEthicalAdversariesMitigating2021}.
Besides, we also compared with the representative algorithms of the other two kinds of fixing algorithms, reweighing in pre-processing~\cite{kamiranDataPreprocessingTechniques2012} and Reject Option Classification in post-processing~\cite{kamiranDecisionTheoryDiscriminationAware2012}.

We aim to answer the following research questions through our experiments:

\noindent \textit{RQ1: How effective is our algorithm in fixing bias model?}

\noindent \textit{RQ2: How efficient is our algorithm in fixing bias model?}

\noindent \textit{RQ3: How parameters affect model performance?} 

\noindent \textit{RQ4: How our algorithm perform in image datasets?} 

\subsection{Effectiveness of \sys}\label{sec:effect}

\noindent \textbf{Experiment Design:}
To evaluate the effectiveness of \sys, we test the following models: the naive baseline model, models fixed by FAD, by Ethical Adversaries, and by \sys. 
We also compared \sys with two popular pre-/post-processing method, reweighing and reject option classification~(ROC).
Due to the randomness in these experiments, we ran the training 10 times to ensure the reliability of results and enforced these fixing algorithms to share the same original training dataset. 
To measure the effectiveness of \sys, we compare the performance between \sys and the other algorithms in terms of both utility and fairness.  
To demonstrate the effectiveness of the three components of \sys~(i.e. neural network slicing, sample clustering and selective training), we conducted a detailed comparison between our algorithm and other popular works.

\noindent \textbf{Results:}
The details of the comparison results are presented in  \autoref{tab:effective}. The first column lists the three datasets. The second column shows the different algorithm. 
The third column lists the utility, and the remaining columns list the fairness criteria. The model utilities are evaluated by binary classification accuracy (Acc), and the fairness performance are measured by demographic parity (DP), demographic parity ratio (DPR), and Equal opportunity (EO). The best results are shown in bold.

\noindent \textbf{Analysis:}
The experimental results demonstrate the effectiveness of our algorithm. Firstly, \sys can effectively fix the fairness bias of all models trained on different datasets. Secondly, \sys achieves the highest utility among all models with fairness constraints, and even surpasses the naive model on COMPAS and Credit.

 \autoref{tab:effective} shows the fairness improvement of naive models on Census, Credit and COMPAS, respectively. \sys improves DPR by 98.47\%, 157.23\%, and 3895.23\%, mitigates fairness problem by 69.69\%, 21.12\% and 38.95\% in terms of EO, and 74.68\%, 2.08\%, 96.19\% in terms of DP. Compared with the other algorithms, \sys achieves the best fairness performance on Census and COMPAS. However, the EO and DP results of \sys on Credit is not satisfactory. After our careful analysis, we found that our neuron network slicing is not fully functional since Credit only has 600 instances. Thus, how to improve the utility of models training on such small datasets will be one of our future works.

Besides, \autoref{tab:effective} demonstrates that \sys has little impact on model utility after a successful fairness fixing, and even has the advantage of increasing accuracy by fixing fairness problems. The average utility of \sys is the highest among all models with fairness constraints, which exceeds ROC by 27.9\%, Reweighing by 17.5\% , Ethical Adversaries by 3.85\% and FAD by 27.22\%, and even surpasses the naive model on the German Credit and COMPAS datasets. The detailed average accuracy change is -0.83\%, 1.36\%, and 28.66\%. We found that it is mainly because \sys improves the utility by mitigating the overfitting problem in model training procedures, and the size of Census dataset is relatively large, so its overfitting problem is unobvious.

In summary, \sys can effectively fix the bias training procedures, and has little impact on the model utility while improving the fairness performance significantly. 
Then we prove the effectiveness of each step in \sys separately.

\subsubsection{Effectiveness of neuron network slicing}

\begin{figure}[h]
    \centering
    \includegraphics[width=0.95\linewidth]{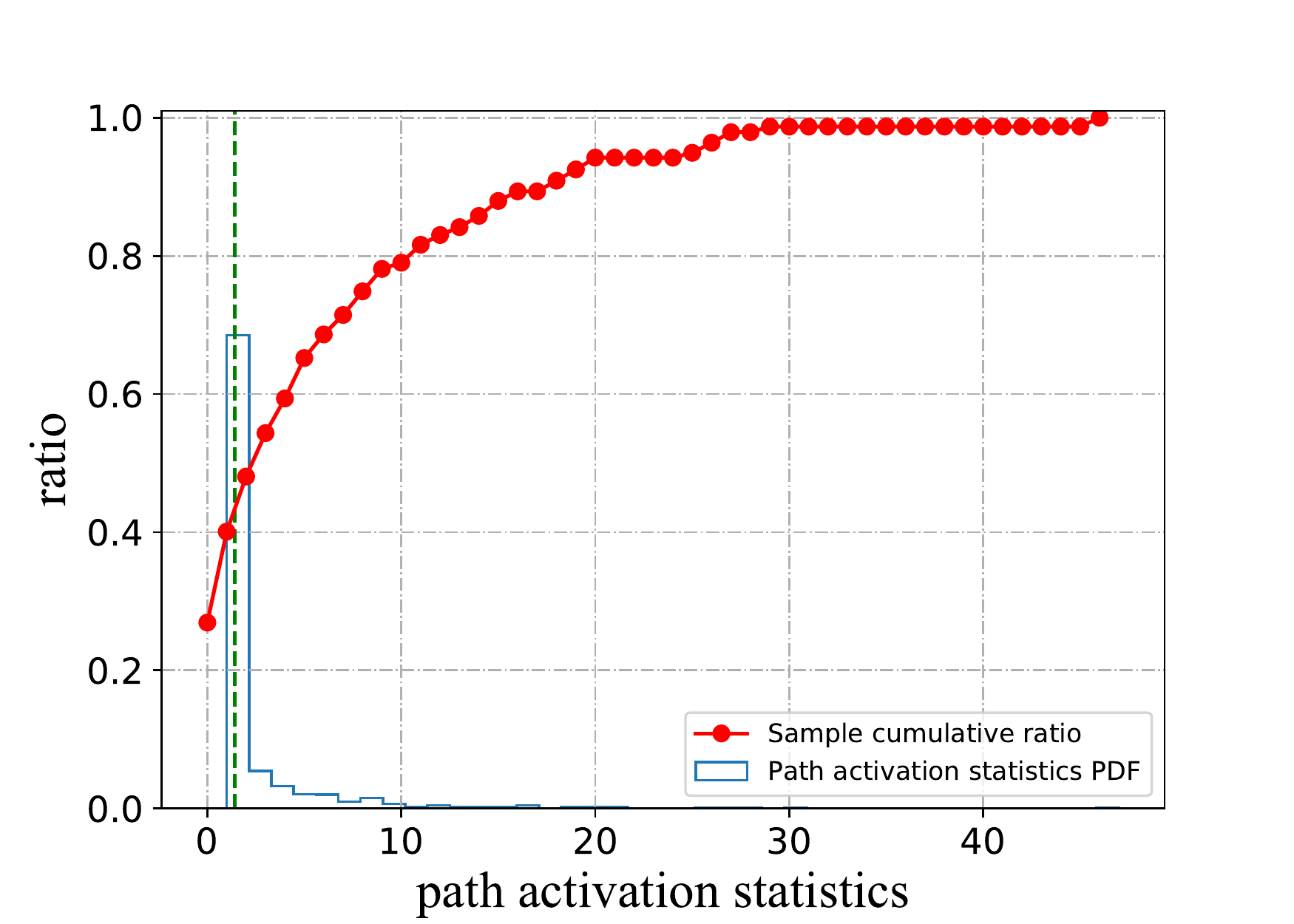}
    \caption{Result of neuron network slicing. The blue step line presents the probability density function of path activation statistics, the red line presents the accumulation of sample ratio, and the green line presents the threshold.}
    \label{fig:detection}
  \end{figure}

\autoref{fig:detection} shows a example of the distribution of abnormal paths. 
Here, the maximum of path activation statistics is 47, and we assume \(\theta=0.03\), so the threshold is 1.41, as the green line shows. 
We can see that most of non-zero paths are concentrated near 1, but the proportion of their corresponding samples is not high.
These paths are the abnormal paths detected by our approach.

\subsubsection{Effectiveness of sample clustering}

To demonstrate the effectiveness of sample clustering, we compare the fixing performance between our sample separation and the random clustering methods. 
We set the number of randomly-obtained abnormal samples to be the same as that of \sys.

\autoref{tab:effective_separation} reports the performance of different clustering methods. With our method, the average accuracy is improved by 6.68\%, and the fairness performance has also been greatly improved, which are 96.19\%, 97.67\% and 38.95\% of DP, DPR and EO, respectively.

\begin{table}[]
    \centering
    \caption{Random clustering vs. our clustering}
    \label{tab:effective_separation}
    \begin{tabular}{@{}lrrrr@{}}
    \toprule
    \multicolumn{1}{c}{Method} & \multicolumn{1}{c}{Acc}   & \multicolumn{1}{c}{DP}    & \multicolumn{1}{c}{DPR}  & \multicolumn{1}{c}{EO}    \\ \midrule
    Random & 0.749 & 0.325 & 1.89 & 0.159 \\
    Ours   & 0.799 & 0.013 & 1.02 & 0.058 \\ \bottomrule
    \end{tabular}
\end{table}

\subsubsection{Effectiveness of selective training}

To show the effectiveness of selective training, we provide a comparison among pure dropout, pure ordinary and selective training.

\begin{table}[]
    \centering
    \caption{Comparison among dropout, ordinary and selective training.}
    \label{tab:effective_dropout}
    \begin{tabular}{@{}lrrrr@{}}
    \toprule
    Training approach & \multicolumn{1}{c}{Acc}   & \multicolumn{1}{c}{DP}    & \multicolumn{1}{c}{DPR}   & \multicolumn{1}{c}{EO}    \\ \midrule
    Ordinary          & 0.575 & 0.733 & 0.183 & 0.683 \\
    Dropout           & 0.621 & 0.341 & 1.860 & 0.095 \\
    Selective   & 0.799 & 0.013 & 1.021 & 0.058 \\ \bottomrule
    \end{tabular}
\end{table}

\autoref{tab:effective_dropout} presents the results of different training approaches. Selective training surpasses the ordinary training by 38.96\%, 98.22\%, 97.43\% and 91.50\%, while surpassing the pure dropout training by 22.27\%, 96.19\%, 97.55\% and 38.95\% in Acc, DP, DPR and EO, respectively.
It confirms that the selective training in \sys can achieve high accuracy and fairness.

\begin{table}[]
        \centering
        \caption{Results on the three datasets. Best results are in bold.}
        \label{tab:effective}
        \begin{tabular}{@{}cccccc@{}}
                \toprule
                Dataset                 & Model               & Acc            & DP             & EO             & DPR            \\ \midrule
                \multirow{4}{*}{Census} & Naive model         & 0.839          & 0.079          & 0.102          & 0.609          \\ \cmidrule(l){2-6} 
                                        & ROC                 & 0.597          & 0.044          & 0.051          & 0.773          \\
                                        & Reweighing                 & 0.719          & 0.059          & 0.0141          & 1.497          \\
                                        & FAD                 & 0.612          & 0.059          & 0.061          & 0.518          \\
                                        & Ethical Adversaries & 0.814          & 0.031          & 0.179          & 0.784          \\
                                        & \sys              & \textbf{0.832} & \textbf{0.020} & \textbf{0.031} & \textbf{0.869} \\ \midrule
                \multirow{4}{*}{Credit} & Naive model         & 0.734          & 0.048          & 0.142          & 0.407          \\ \cmidrule(l){2-6} 
                                        & ROC                 & 0.646          & 0.041          & 0.073          & 1.273          \\
                                        & Reweighing                 & 0.632          & 0.067          & 0.066          & 0.828          \\
                                        & FAD                 & 0.710          & \textbf{0.000} & \textbf{0.000} & inf            \\
                                        & Ethical Adversaries & 0.715          & 0.041          & 0.031          & 2.442          \\
                                        & \sys              & \textbf{0.744} & 0.047          & 0.112          & \textbf{0.834} \\ \midrule
                \multirow{4}{*}{COMPAS} & Naive model         & 0.621          & 0.341          & 0.095          & 1.860           \\ \cmidrule(l){2-6} 
                                        & ROC                 & 0.618          & 0.083          & 0.069          & 0.890          \\
                                        & Reweighing                 & 0.671          & 0.193          & 0.176          & 1.406          \\
                                        & FAD                 & 0.567          & 0.057          & 0.114          & 0.926          \\
                                        & Ethical Adversaries & 0.759          & 0.095          & 0.095          & 1.203          \\
                                        & \sys              & \textbf{0.799} & \textbf{0.013} & \textbf{0.058} & \textbf{1.021} \\ \bottomrule
                \end{tabular}
\end{table}

\subsection{Efficiency of \sys}\label{sec:efficient}

\noindent \textbf{Experiment Design:}
To evaluate the efficiency of \sys, we measured the time usage of ordinary training, Ethical Adversaries and \sys training on all three datasets. We performed 10 trials which uses random training/test data splitting, naive model training, hyperparameter tuning and model repairing~(for Ethical Adversaries and \sys) and computed the average overhead to avoid randomness.
\autoref{tab:efficient} presents how much time it takes to complete its fixing for each method. For Ethical Adversaries, it shows the time for per iteration in 50 iterations. For \sys, it shows the time usage per trial. We also recorded the time usage of each step in \sys. Results and analysis are presented below.

\noindent \textbf{Results:}
\autoref{tab:efficient} shows the comparison of time usage among ordinary training, Ethical Adversaries and \sys training.
The first column lists the three datasets and the remaining columns show the different training methods. On average, \sys takes 5.39 times of ordinary training and 55.49\% of Ethical Adversaries.

\autoref{tab:step_efficient} reports the time costs of each step. The first column also lists the three datasets and the remaining columns show the time costs of hyperparameters selection, neuron network slicing, sample clustering and selective training respectively.

\noindent \textbf{Analysis:}
For ordinary training, the runtime overhead all comes from the training procedure, but for \sys, the hyperparameters tuning accounts for a larger ratio of the total time usage, as shown in \autoref{tab:step_efficient}.
So \sys takes only less than twice of the time usage of ordinary training on large datasets like Census, but on small datasets like the German Credit dataset, it takes relatively a long time.
If \sys tries more times, the average time will be reduced because the hyperparameters tuning is only conducted once.

Overall, \sys is more efficient than Ethical Adversaries in fixing models, with an average speedup of 180\%.

\begin{table}[]
        \centering
        \caption{Time to train a model.}
        \label{tab:efficient}
        \begin{tabular}{@{}lrrr@{}}
        \toprule
        Dataset & Naive  & EA (/iteration) & \sys(/trial) \\ 
        \midrule
        Census  & 115.74s & 1439.96s           & 254.41s            \\ 
        Credit  & 3.07s  & 33.24s             & 31.49s             \\ 
        COMPAS  & 11.92s  & 81.93s             & 44.31s             \\ 
        \bottomrule
        \end{tabular}
\end{table}

\begin{table}[]
    \centering
    \caption{Time used in each step.}
    \label{tab:step_efficient}
    \begin{tabular}{@{}lrrrr@{}}
    \toprule
    Dataset & Para selection & Slicing & Clustering & Training \\ \midrule
    Census  & 115.41s         & 25.37s        & 43.70s            & 74.37s          \\
    Credit  & 30.98s          & 0.20s         & 6.73e-4s          & 0.30s           \\
    COMPAS  & 40.76s          & 2.09s         & 0.06s             & 1.40s           \\ \bottomrule
    \end{tabular}
\end{table}

\subsection{Effects of Configurable Hyperparameters}\label{sec:param}

\sys leverages two configurable hyperparameters, $\theta$ and $\gamma$, to fix fairness problems. 
The hyperparameters $\gamma$ represents the threshold of neuron activation, and its value affects the complexity of the path. 
As its value decreases, more neurons and synapses are included in the path, resulting in a more complex path. 
And $\theta$ represents the threshold of neuron network slicing. 
The lower is the $\theta$, the fewer paths are classified as abnormal.

We conduct a comparison experiment of these hyperparameters. $\theta$ varies between the interval $[10^{-4},1]$ and $\gamma$ varies between the interval $[0.5,1]$. 
Note that we use logarithmic coordinates for $\theta$ since its value is sampled proportionally.

\begin{figure*}[]
        \centering
        \footnotesize
        \begin{subfigure}[t]{0.325\textwidth}
                \centering     
                \footnotesize
                \includegraphics[width=\textwidth]{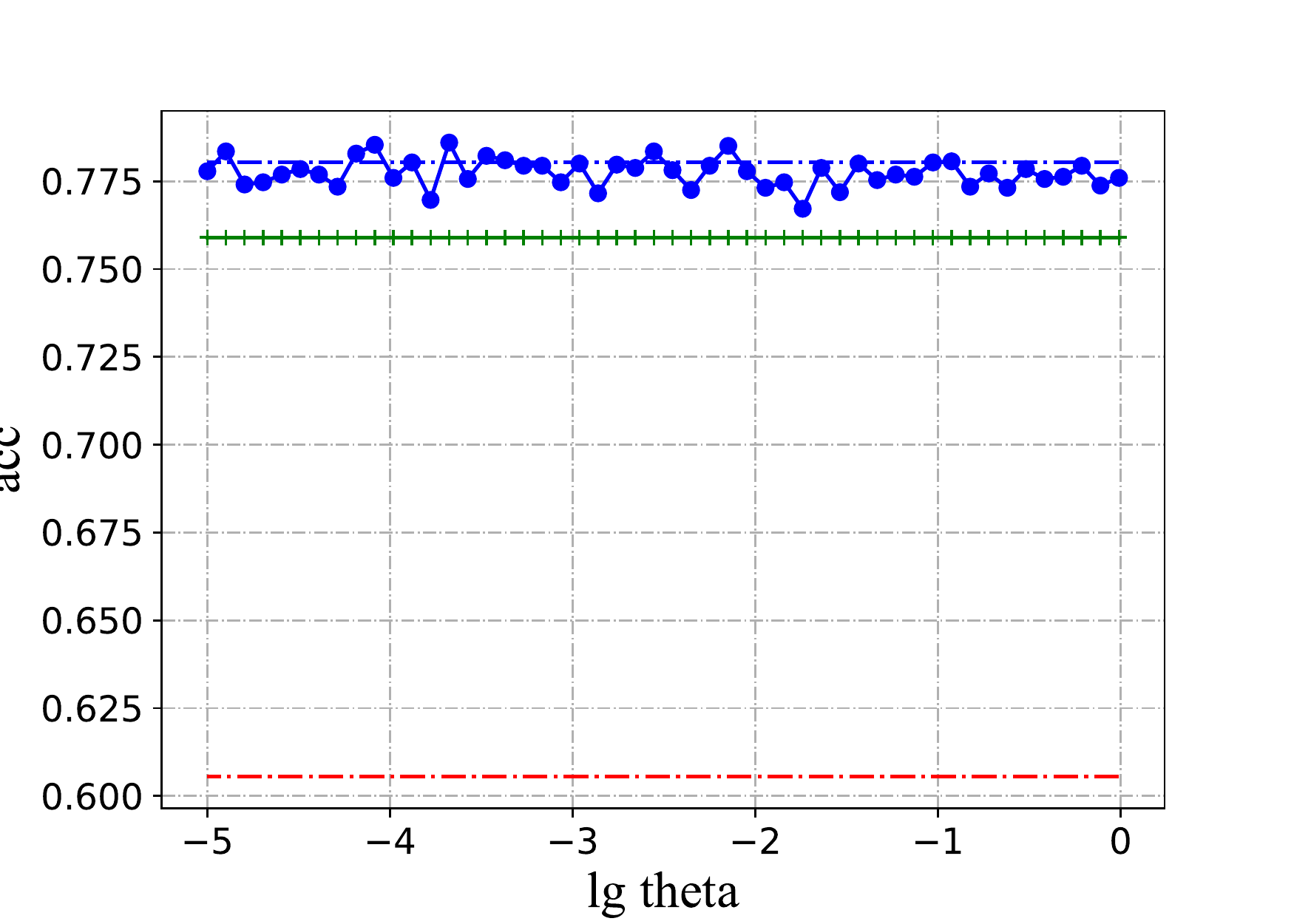}
                       \caption{$\theta$-accuracy}\label{f:theta_acc}
            \end{subfigure}
            \hfill
            \begin{subfigure}[t]{0.325\textwidth}
                \centering
                       \footnotesize
                       \includegraphics[width=\textwidth]{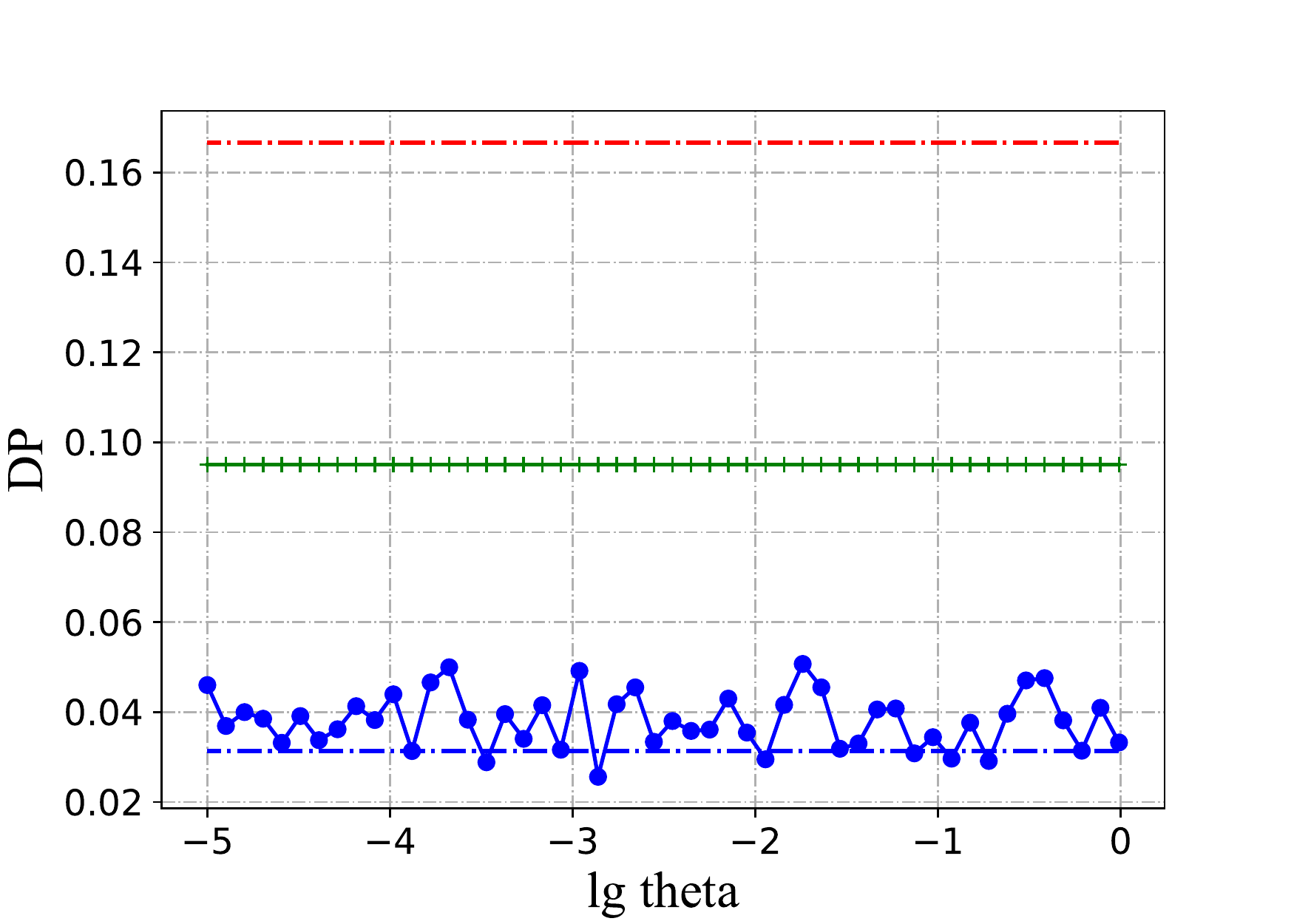}
                       \caption{$\theta$-DP}\label{f:theta_dp}
            \end{subfigure}
            \begin{subfigure}[t]{0.325\textwidth}
                \centering     
                \footnotesize
                \includegraphics[width=\textwidth]{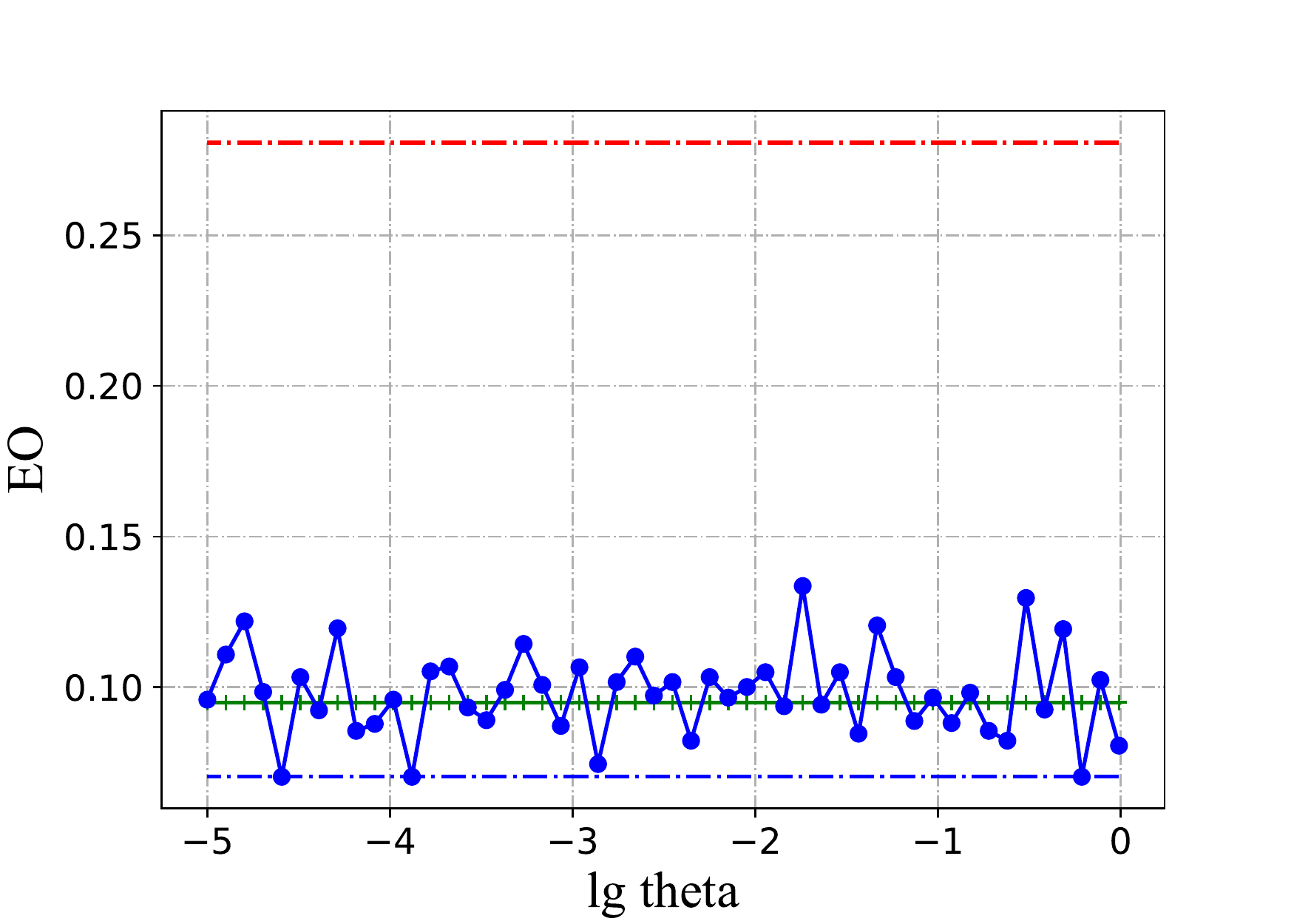}
                \caption{$\theta$-EO}\label{f:theta_eo}
            \end{subfigure}
            
            \begin{subfigure}[t]{0.325\textwidth}
                \centering     
                \footnotesize
                \includegraphics[width=\textwidth]{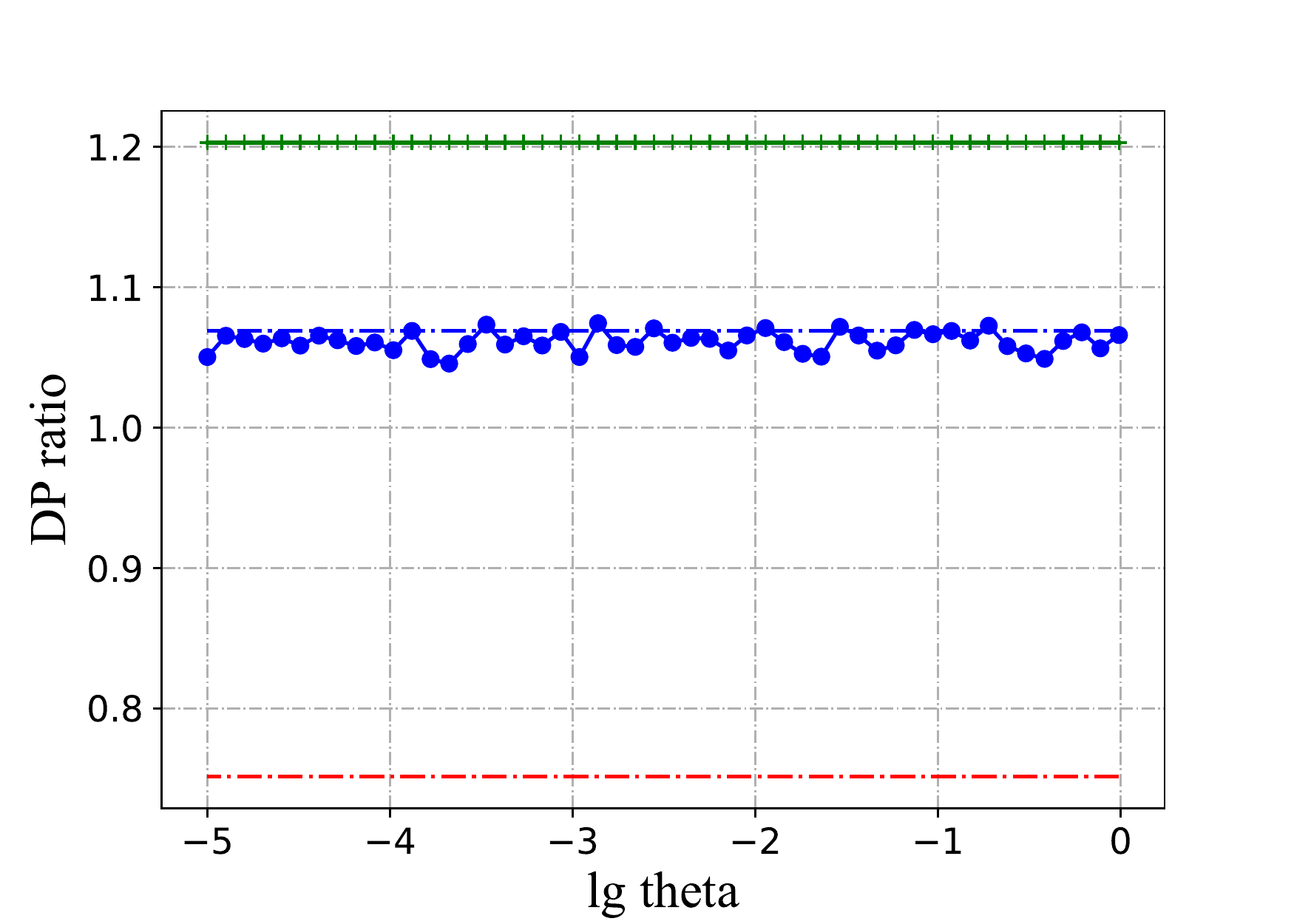}
                       \caption{$\theta$-DPR}\label{f:theta_dpr}
            \end{subfigure}
            \hfill
        \begin{subfigure}[t]{0.325\textwidth}
            \centering     
            \footnotesize
            \includegraphics[width=\textwidth]{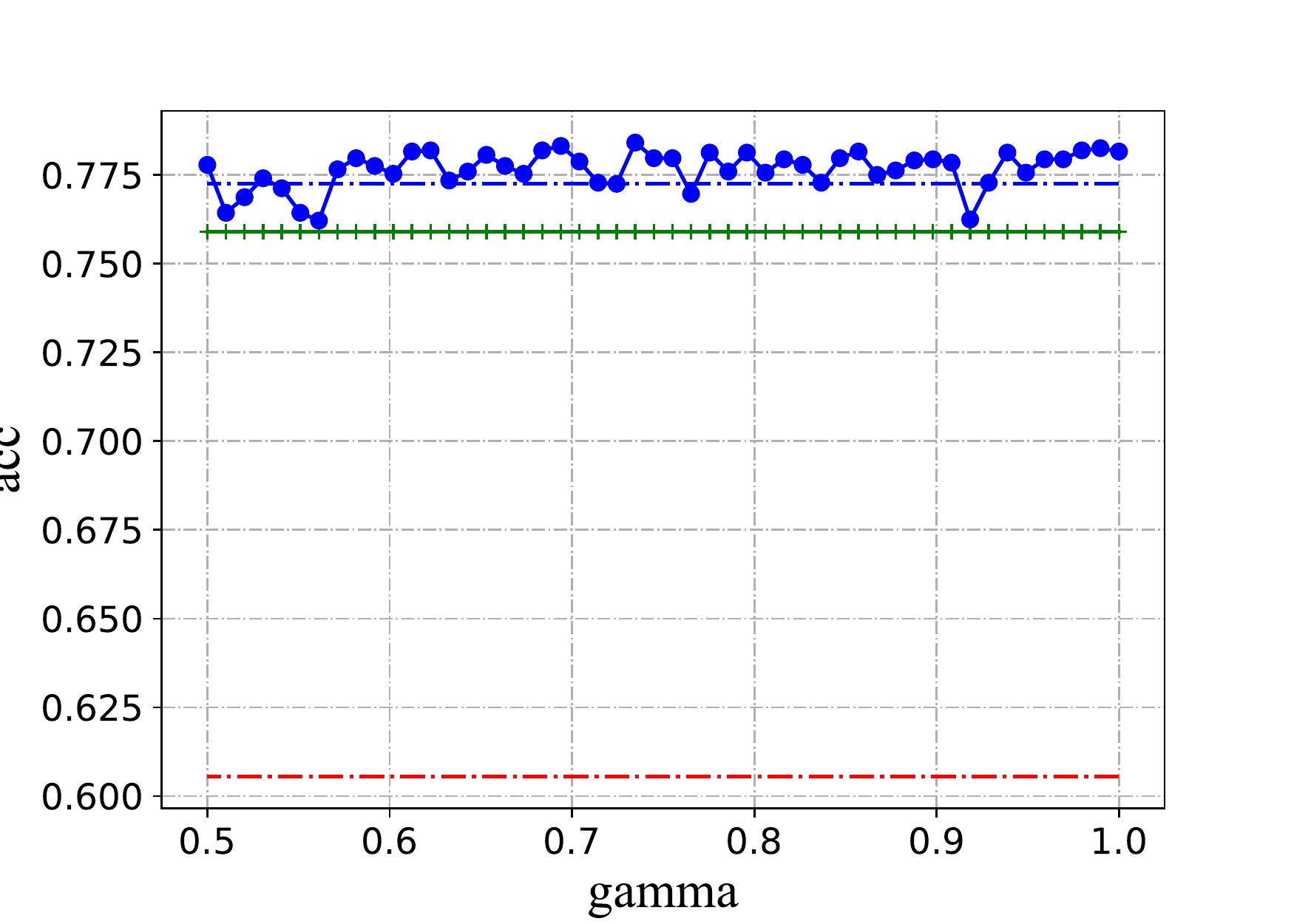}
            \caption{$\gamma$-accuracy}\label{f:gamma_acc}
        \end{subfigure}
        \hfill
        \begin{subfigure}[t]{0.325\textwidth}
            \centering     
            \footnotesize
            \includegraphics[width=\textwidth]{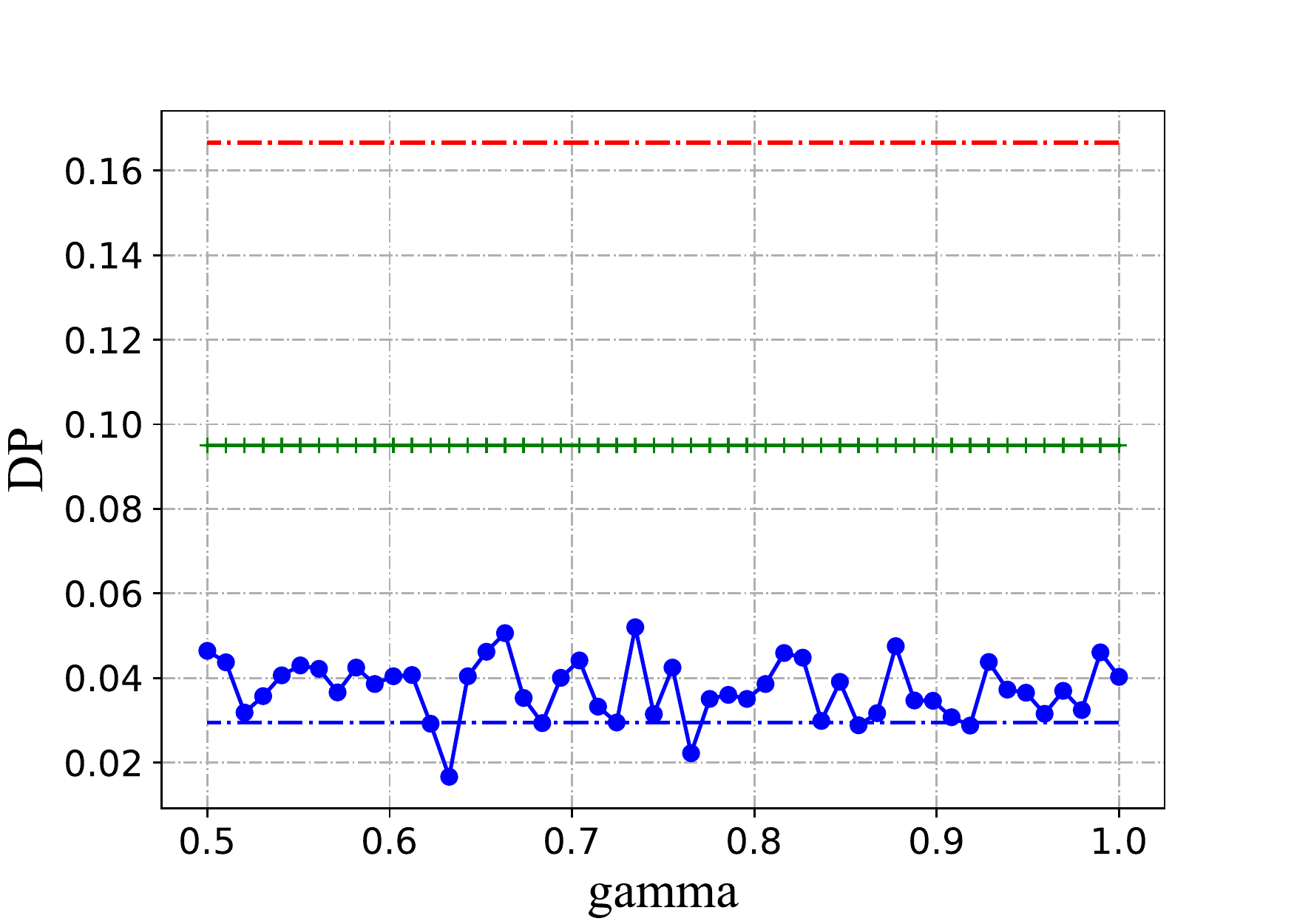}
                \caption{$\gamma$-DP}\label{f:gamma_dp}
        \end{subfigure}
        
        \begin{subfigure}[t]{0.325\textwidth}
            \centering
                \footnotesize
                \includegraphics[width=\textwidth]{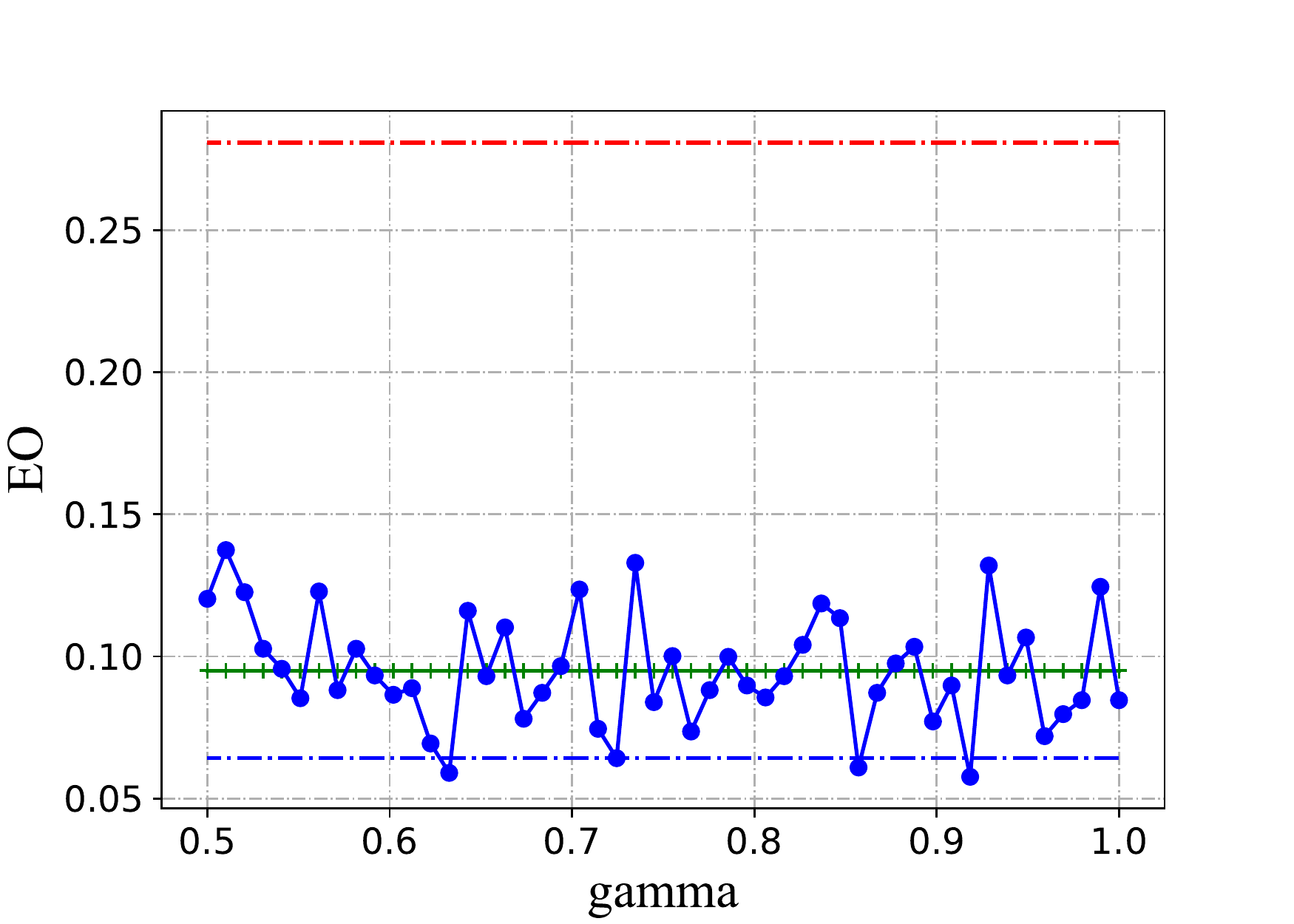}
                \caption{$\gamma$-EO}\label{f:gamma_eo}
        \end{subfigure}
        \hfill
        \begin{subfigure}[t]{0.325\textwidth}
                \centering     
                \footnotesize
                \includegraphics[width=\textwidth]{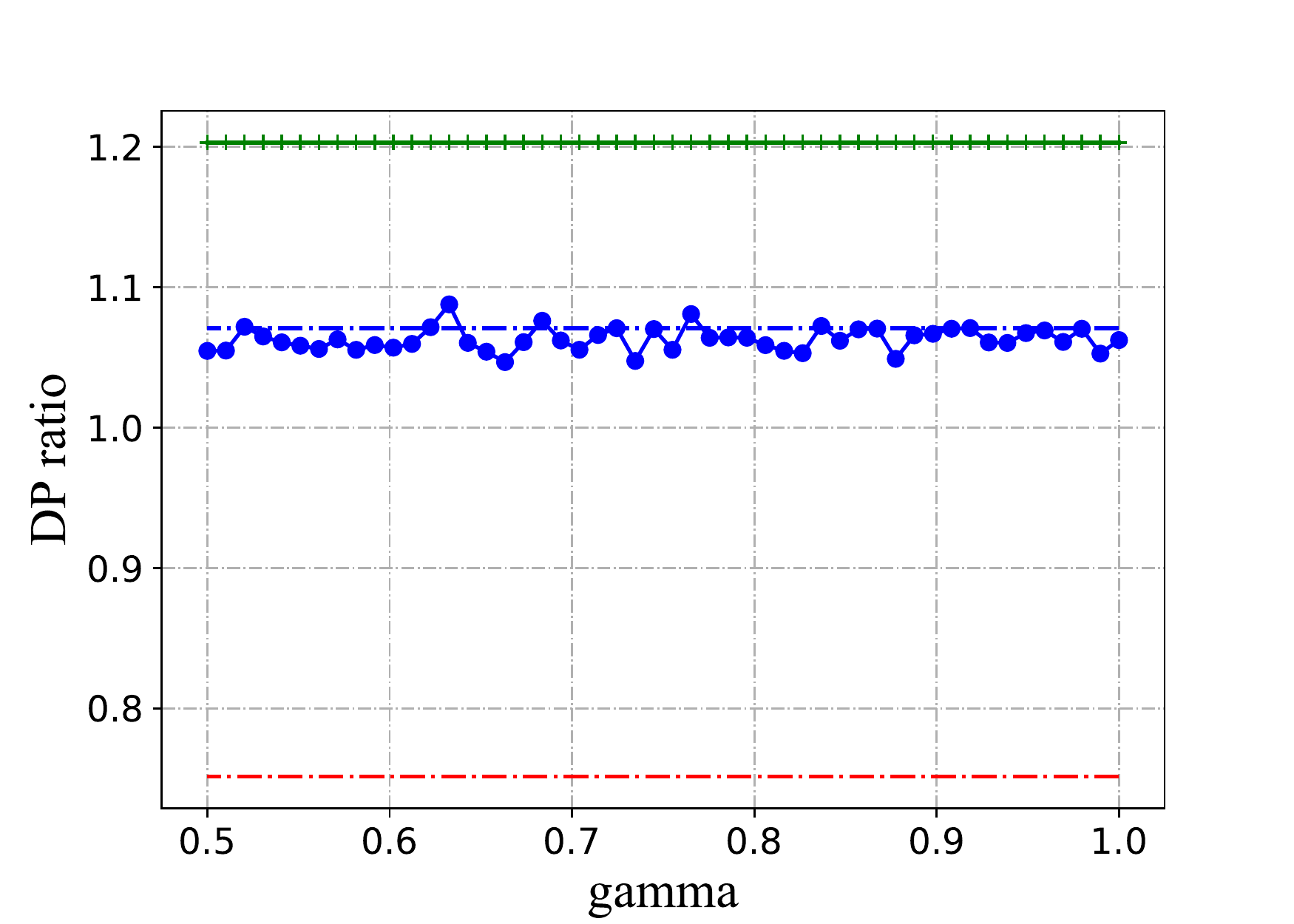}
                \caption{$\gamma$-DPR}\label{f:gamma_dpr}
            \end{subfigure}\
        \hfill
        \begin{subfigure}[t]{0.325\textwidth}
            \centering
                    \footnotesize
                    \includegraphics[trim={0 0 5 0},clip,width=0.5\textwidth]{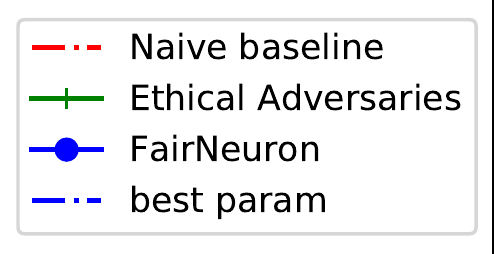}
        \end{subfigure}

        \caption{Effect of hyperparameters $\theta$ and $\gamma$. $\theta$ is sampled proportionally, so we take the logarithm of $\theta$ as x axis.}
        \label{fig:hyperparameters}
   \end{figure*}

\autoref{fig:hyperparameters} shows how hyperparameters assignments will affect the performance. 
Based on our results in comparison with the naive model and Ethical Adversaries, we can conclude that our algorithm does perform better on this task and is not sensitive to hyperparameters assignments except for EO~(\autoref{fig:hyperparameters}(c) \& (g)). 
By increasing the weight of EO in hyperparameter tuning loss function, we can constrain its fluctuations.

\subsection{Performance on Image Datasets}\label{sec:image_dataset}
\noindent \textbf{Experiment Design:}
We also explored the possibility of using our method in fixing models on image datasets, which has not done by baseline methods due to their inefficiency.
In our experiment, we leverage a 4-layer fully-connected NN trained on MNIST~\cite{MNISTHandwrittenDigit} and ResNet-18~\cite{heDeepResidualLearning2016} trained on CIFAR-10~\cite{CIFAR10CIFAR100Datasets}, compare \sys with the naive model and random dropout.
We use Class-wise Variance~(CV) and Maximum Class-wise Discrepancy~(MCD) as fairness metrics.

\noindent \textbf{Results:}
\autoref{tab:imagetable} summarizes our results. 
The first column lists the datasets.
The second column shows the different model, and the remaining columns list the performance.
The best results are shown in bold.
As we can see from the table, \sys can effectively improve the model fairness by 20\% for MCD and 80\% for CV.
We discuss it further in \autoref{sec:conclusion}.

\begin{table}[]
    \centering
    \caption{Results on image datasets. Best results are in bold.}
    \label{tab:imagetable}
    \begin{tabular}{llrrr}
    \toprule
    \multicolumn{1}{c}{Dataset}               & \multicolumn{1}{c}{Model}               & \multicolumn{1}{c}{Acc}          & \multicolumn{1}{c}{CV}             & \multicolumn{1}{c}{MCD}          \\ \midrule
    \multirow{3}{*}{MNIST} & Naive model         & 0.957          & 6.66e-5          & 0.057          \\
                           & Random dropout      & 0.949          & 3.58e-5          & 0.052          \\
                           & \sys & \textbf{0.961} & \textbf{9.54e-6} & \textbf{0.046} \\
    \midrule
    \multirow{3}{*}{CIFAR-10} & Naive model         & \textbf{0.814} & 6.04e-4          & 0.236          \\
                           & Random dropout      & 0.798          & 8.55e-4          & 0.464          \\
                           & \sys & 0.808          & \textbf{1.16e-4} & \textbf{0.187} \\
    \bottomrule
    \end{tabular}
    \end{table}



\section{Related Work}\label{sec:rw}

\noindent
\textbf{Neural Network Slicing}.
Path analysis or dataflow analysis~\cite{ammonsImprovingDataflowAnalysis} is a fundamental technique in traditional software engineering tasks like testing, debugging and optimization. 
It offers a window to study program's dynamic behavior.    
In recent years, with the development of AI security, especially adversarial attack and defense, conflict path detection has been used for interpretability.
Wang et al.~\cite{wangInterpretNeuralNetworks2018} proposed a method to interpret neural networks by extracting the critical data routing paths~(CDRPs), and they demonstrated its effectiveness on adversarial sample detection problem. 
Qiu et al.~\cite{qiuAdversarialDefenseNetwork2019} treat a neural network as a dataflow graph, which can be applied the profiling technique to extract its execution path. 
Zhang et al.~\cite{zhangDynamicSlicingDeep2020} apply the dynamic slicing on deep neural networks.

\noindent
\textbf{Fairness of ML}.
With the increasing use of automated decision-making approaches and systems, fairness considerations in ML have gained significant attention. 
Researchers found many fairness problems with high social impact, such as standardized tests in higher education~\cite{clearyTestBiasValidity1966}, employment~\cite{guionEmploymentTestsDiscriminatory1966, raghavanMitigatingBiasAlgorithmic2020, vandenbroekHiringAlgorithmsEthnography2019}, and re-offence judgement~\cite{oneilWeaponsMathDestruction2016, brennanEmergenceMachineLearning2013, berkFairnessCriminalJustice2021, berkAccuracyFairnessJuvenile2019}. 
Besides, governments~(e.g. the EU~\cite{voigtEuGeneralData2017} and the US~\cite{presidentBigDataSeizing2014, presidentBigDataReport2016}), organizations~\cite{markhamEthicalDecisionmakingInternet2012}, and the media have called for more societal accountability and social understanding of ML.

To address the concern above, numerous fairness notions are proposed.
In high level, these fairness notions can be split into three categories: (i) individual fairness, which requires that similar individuals should be treated similarly~\cite{dworkFairnessAwareness2012, zemelLearningFairRepresentations, lahotiOperationalizingIndividualFairness2019}; (ii) group fairness, which concerns about whether subpopulation with different sensitive characteristics are treated equally~\cite{feldmanCertifyingRemovingDisparate2015,hardtEqualityOpportunitySupervised2016, zhangAchievingNonDiscriminationData2017}; (iii) Max-Min fairness, which try to improve the per-group fairness~\cite{hashimotoFairnessDemographicsRepeated2018,lahotiFairnessDemographicsAdversarially2020,zhangFairnessMultiagentSequential2014}.

Fairness testing is also an important research direction, and its approaches mostly based on generation techniques. 
THEMIS~\cite{angellThemisAutomaticallyTesting2018} considers group fairness using causal analysis and uses random test generation to evaluate fairness.
AEQUITAS inherits and improves THEMIS, and focuses on the individual discriminatory instances generation~\cite{udeshiAutomatedDirectedFairness2018}. 
Later, ADF combines global search and local search to systematically search the input space with the guidance of gradient~\cite{zhangWhiteboxFairnessTesting2020}. 
Symbolic Generation~(SG) integrates symbolic execution and local model explanation techniques to craft individual discriminatory instances~\cite{agarwalAutomatedTestGeneration2018}.

The ML model needs to be repaired after the fairness problem is found.
These approaches can be generally split into three categories: (i) Pre-processing approaches, which fix the training data to reduce the latent discrimination in dataset. For example, the bias could be mitigated by correcting labels~\cite{kamiranClassifyingDiscriminating2009, zhangAchievingNonDiscriminationData2017}, revising attributes~\cite{feldmanCertifyingRemovingDisparate2015, kamiranDataPreprocessingTechniques2012}, generating non-discrimination data~\cite{sattigeriFairnessGANGenerating2019, xuFairganFairnessawareGenerative2018}, and obtaining fair data representations~\cite{beutelDataDecisionsTheoretical2017}. 
(ii) In-processing approaches, which revise the training of the bias model to achieve fairness~\cite{zafarFairnessConstraintsMechanisms2017, zhangMitigatingUnwantedBiases2018}. 
More specifically, these approaches apply fairness constraints~\cite{zafarFairnessConstraintsMechanisms2017,dworkFairnessAwareness2012}, propose an objective function considering the fairness of prediction~\cite{zhangMitigatingUnwantedBiases2018}, or design a new training frameworks~\cite{xuFairganFairnessawareGenerative2018,adelOneNetworkAdversarialFairness2019}. 
(iii) post-processing approaches, which directly change the predictive labels of bias models' output to obtain fairness~\cite{hardtEqualityOpportunitySupervised2016, pleissFairnessCalibration}.

\section{Conclusion}\label{sec:conclusion}

In this paper, we proposed a lightweight algorithm \sys to effectively fixing fairness problems for deep neural network through path analysis. Our algorithm combines a path analysis procedure and a dropout procedure to systematically improve model performance. \sys searches bias instances with the guidance of path analysis and mitigates fairness problems by dropout training.
Our evaluation results show that \sys has significantly better performance both in terms of effectively and efficiently in fixing bias models.
For CNN model, we can only perform \sys on the last full-connected layer, so its performance is not ideal. 
We will improve \sys on CNN in the future.
\section{Acknowledgement}\label{sec:ack}

We thank the anonymous reviewers for their constructive comments. 
This research was partially supported by National Key R\&D Program (2020YFB1406900), National Natural Science Foundation of China (U21B2018, 62161160337, 61822309, U20B2049, 61773310, U1736205, 
61802166) and Shaanxi Province Key Industry Innovation Program (2021ZDLGY01-02). 
Chao Shen is the corresponding author. 
The views, opinions and/or findings expressed are only those of the authors.

\bibliographystyle{ACM-Reference-Format}
\bibliography{FIXATE}


\begin{thebibliography}{73}


\ifx \showCODEN    \undefined \def \showCODEN     #1{\unskip}     \fi
\ifx \showDOI      \undefined \def \showDOI       #1{#1}\fi
\ifx \showISBNx    \undefined \def \showISBNx     #1{\unskip}     \fi
\ifx \showISBNxiii \undefined \def \showISBNxiii  #1{\unskip}     \fi
\ifx \showISSN     \undefined \def \showISSN      #1{\unskip}     \fi
\ifx \showLCCN     \undefined \def \showLCCN      #1{\unskip}     \fi
\ifx \shownote     \undefined \def \shownote      #1{#1}          \fi
\ifx \showarticletitle \undefined \def \showarticletitle #1{#1}   \fi
\ifx \showURL      \undefined \def \showURL       {\relax}        \fi
\providecommand\bibfield[2]{#2}
\providecommand\bibinfo[2]{#2}
\providecommand\natexlab[1]{#1}
\providecommand\showeprint[2][]{arXiv:#2}

\bibitem[\protect\citeauthoryear{??}{CIF}{[n.d.]}]%
        {CIFAR10CIFAR100Datasets}
 \bibinfo{year}{[n.d.]}\natexlab{}.
\newblock \bibinfo{title}{{CIFAR}-10 and {CIFAR}-100 datasets}.
\newblock
\newblock
\urldef\tempurl%
\url{https://www.cs.toronto.edu/~kriz/cifar.html}
\showURL{%
\tempurl}


\bibitem[\protect\citeauthoryear{??}{Fir}{[n.d.]}]%
        {FirstInternationalBeauty}
 \bibinfo{year}{[n.d.]}\natexlab{}.
\newblock \bibinfo{title}{The {First} {International} {Beauty} {Contest}
  {Judged} by {Artificial} {Intelligence}}.
\newblock
\newblock
\urldef\tempurl%
\url{http://beauty.ai}
\showURL{%
\tempurl}


\bibitem[\protect\citeauthoryear{??}{Mac}{[n.d.]}]%
        {MachineBiasProPublica}
 \bibinfo{year}{[n.d.]}\natexlab{}.
\newblock \bibinfo{title}{Machine {Bias} — {ProPublica}}.
\newblock
\newblock
\urldef\tempurl%
\url{https://www.propublica.org/article/machine-bias-risk-assessments-in-criminal-sentencing}
\showURL{%
\tempurl}


\bibitem[\protect\citeauthoryear{??}{Mic}{[n.d.]}]%
        {MicrosoftNeoNaziSexbot}
 \bibinfo{year}{[n.d.]}\natexlab{}.
\newblock \bibinfo{title}{Microsoft’s neo-{Nazi} sexbot was a great lesson
  for makers of {AI} assistants}.
\newblock
\newblock
\urldef\tempurl%
\url{https://www.technologyreview.com/2018/03/27/144290/microsofts-neo-nazi-sexbot-was-a-great-lesson-for-makers-of-ai-assistants/}
\showURL{%
\tempurl}


\bibitem[\protect\citeauthoryear{??}{MNI}{[n.d.]}]%
        {MNISTHandwrittenDigit}
 \bibinfo{year}{[n.d.]}\natexlab{}.
\newblock \bibinfo{title}{{MNIST} handwritten digit database, {Yann} {LeCun},
  {Corinna} {Cortes} and {Chris} {Burges}}.
\newblock
\newblock
\urldef\tempurl%
\url{http://yann.lecun.com/exdb/mnist/}
\showURL{%
\tempurl}


\bibitem[\protect\citeauthoryear{??}{Tun}{[n.d.]}]%
        {TuneScalableHyperparameter}
 \bibinfo{year}{[n.d.]}\natexlab{}.
\newblock \bibinfo{title}{Tune: {Scalable} {Hyperparameter} {Tuning} — {Ray}
  v1.9.0}.
\newblock
\newblock
\urldef\tempurl%
\url{https://docs.ray.io/en/latest/tune/index.html}
\showURL{%
\tempurl}


\bibitem[\protect\citeauthoryear{??}{Bea}{2016}]%
        {BeautyContestWas2016}
 \bibinfo{year}{2016}\natexlab{}.
\newblock \bibinfo{title}{A beauty contest was judged by {AI} and the robots
  didn't like dark skin}.
\newblock
\newblock
\urldef\tempurl%
\url{http://www.theguardian.com/technology/2016/sep/08/artificial-intelligence-beauty-contest-doesnt-like-black-people}
\showURL{%
\tempurl}
\newblock
\shownote{Section: Technology.}


\bibitem[\protect\citeauthoryear{??}{NAB}{2020}]%
        {NABTurnsAI2020}
 \bibinfo{year}{2020}\natexlab{}.
\newblock \bibinfo{title}{{NAB} turns to {AI} to decide on small business
  loans}.
\newblock
\newblock
\urldef\tempurl%
\url{https://www.afr.com/companies/financial-services/nab-turns-to-artificial-intelligence-to-assess-small-business-loans-20201204-p56kmk}
\showURL{%
\tempurl}
\newblock
\shownote{Section: financialservices.}


\bibitem[\protect\citeauthoryear{??}{How}{2021}]%
        {HowAIWill2021}
 \bibinfo{year}{2021}\natexlab{}.
\newblock \bibinfo{title}{How {AI} will change the {HR} industry {\textbar}
  {HRExecutive}.com}.
\newblock
\newblock
\urldef\tempurl%
\url{http://hrexecutive.com/ai-will-make-traditional-hr-extinct-how-to-prepare-for-whats-next/}
\showURL{%
\tempurl}


\bibitem[\protect\citeauthoryear{Abadi, Barham, Chen, Chen, Davis, Dean, Devin,
  Ghemawat, Irving, Isard, Kudlur, Levenberg, Monga, Moore, Murray, Steiner,
  Tucker, Vasudevan, Warden, Wicke, Yu, and Zheng}{Abadi et~al\mbox{.}}{2016}]%
        {abadiTensorFlowSystemLargeScale2016}
\bibfield{author}{\bibinfo{person}{Martin Abadi}, \bibinfo{person}{Paul
  Barham}, \bibinfo{person}{Jianmin Chen}, \bibinfo{person}{Zhifeng Chen},
  \bibinfo{person}{Andy Davis}, \bibinfo{person}{Jeffrey Dean},
  \bibinfo{person}{Matthieu Devin}, \bibinfo{person}{Sanjay Ghemawat},
  \bibinfo{person}{Geoffrey Irving}, \bibinfo{person}{Michael Isard},
  \bibinfo{person}{Manjunath Kudlur}, \bibinfo{person}{Josh Levenberg},
  \bibinfo{person}{Rajat Monga}, \bibinfo{person}{Sherry Moore},
  \bibinfo{person}{Derek~G. Murray}, \bibinfo{person}{Benoit Steiner},
  \bibinfo{person}{Paul Tucker}, \bibinfo{person}{Vijay Vasudevan},
  \bibinfo{person}{Pete Warden}, \bibinfo{person}{Martin Wicke},
  \bibinfo{person}{Yuan Yu}, {and} \bibinfo{person}{Xiaoqiang Zheng}.}
  \bibinfo{year}{2016}\natexlab{}.
\newblock \showarticletitle{{TensorFlow}: {A} {System} for {Large}-{Scale}
  {Machine} {Learning}}. \bibinfo{pages}{265--283}.
\newblock
\showISBNx{978-1-931971-33-1}
\urldef\tempurl%
\url{https://www.usenix.org/conference/osdi16/technical-sessions/presentation/abadi}
\showURL{%
\tempurl}


\bibitem[\protect\citeauthoryear{Adel, Valera, Ghahramani, and Weller}{Adel
  et~al\mbox{.}}{2019}]%
        {adelOneNetworkAdversarialFairness2019}
\bibfield{author}{\bibinfo{person}{Tameem Adel}, \bibinfo{person}{Isabel
  Valera}, \bibinfo{person}{Zoubin Ghahramani}, {and} \bibinfo{person}{Adrian
  Weller}.} \bibinfo{year}{2019}\natexlab{}.
\newblock \showarticletitle{One-{Network} {Adversarial} {Fairness}}.
\newblock \bibinfo{journal}{\emph{Proceedings of the AAAI Conference on
  Artificial Intelligence}}  \bibinfo{volume}{33} (\bibinfo{date}{July}
  \bibinfo{year}{2019}), \bibinfo{pages}{2412--2420}.
\newblock
\showISSN{2374-3468, 2159-5399}
\urldef\tempurl%
\url{https://doi.org/10.1609/aaai.v33i01.33012412}
\showDOI{\tempurl}


\bibitem[\protect\citeauthoryear{Agarwal, Lohia, Nagar, Dey, and Saha}{Agarwal
  et~al\mbox{.}}{2018}]%
        {agarwalAutomatedTestGeneration2018}
\bibfield{author}{\bibinfo{person}{Aniya Agarwal}, \bibinfo{person}{Pranay
  Lohia}, \bibinfo{person}{Seema Nagar}, \bibinfo{person}{Kuntal Dey}, {and}
  \bibinfo{person}{Diptikalyan Saha}.} \bibinfo{year}{2018}\natexlab{}.
\newblock \showarticletitle{Automated test generation to detect individual
  discrimination in {AI} models}.
\newblock \bibinfo{journal}{\emph{arXiv preprint arXiv:1809.03260}}
  (\bibinfo{year}{2018}).
\newblock


\bibitem[\protect\citeauthoryear{Ammons and Larust}{Ammons and
  Larust}{[n.d.]}]%
        {ammonsImprovingDataflowAnalysis}
\bibfield{author}{\bibinfo{person}{Glenn Ammons} {and} \bibinfo{person}{James~R
  Larust}.} \bibinfo{year}{[n.d.]}\natexlab{}.
\newblock \showarticletitle{Improving {Data}-flow {Analysis} with {Path}
  {Profiles}}.
\newblock  (\bibinfo{year}{[n.\,d.]}), \bibinfo{pages}{13}.
\newblock


\bibitem[\protect\citeauthoryear{Angell, Johnson, Brun, and Meliou}{Angell
  et~al\mbox{.}}{2018}]%
        {angellThemisAutomaticallyTesting2018}
\bibfield{author}{\bibinfo{person}{Rico Angell}, \bibinfo{person}{Brittany
  Johnson}, \bibinfo{person}{Yuriy Brun}, {and} \bibinfo{person}{Alexandra
  Meliou}.} \bibinfo{year}{2018}\natexlab{}.
\newblock \showarticletitle{Themis: automatically testing software for
  discrimination}. In \bibinfo{booktitle}{\emph{Proceedings of the 2018 26th
  {ACM} {Joint} {Meeting} on {European} {Software} {Engineering} {Conference}
  and {Symposium} on the {Foundations} of {Software} {Engineering} -
  {ESEC}/{FSE} 2018}}. \bibinfo{publisher}{ACM Press}, \bibinfo{address}{Lake
  Buena Vista, FL, USA}, \bibinfo{pages}{871--875}.
\newblock
\showISBNx{978-1-4503-5573-5}
\urldef\tempurl%
\url{https://doi.org/10.1145/3236024.3264590}
\showDOI{\tempurl}


\bibitem[\protect\citeauthoryear{Barocas and Selbst}{Barocas and
  Selbst}{2016}]%
        {barocasBigDataDisparate2016}
\bibfield{author}{\bibinfo{person}{Solon Barocas} {and}
  \bibinfo{person}{Andrew~D. Selbst}.} \bibinfo{year}{2016}\natexlab{}.
\newblock \showarticletitle{Big data's disparate impact}.
\newblock \bibinfo{journal}{\emph{Calif. L. Rev.}}  \bibinfo{volume}{104}
  (\bibinfo{year}{2016}), \bibinfo{pages}{671}.
\newblock
\newblock
\shownote{Publisher: HeinOnline.}


\bibitem[\protect\citeauthoryear{Berk}{Berk}{2019}]%
        {berkAccuracyFairnessJuvenile2019}
\bibfield{author}{\bibinfo{person}{Richard Berk}.}
  \bibinfo{year}{2019}\natexlab{}.
\newblock \showarticletitle{Accuracy and fairness for juvenile justice risk
  assessments}.
\newblock \bibinfo{journal}{\emph{Journal of Empirical Legal Studies}}
  \bibinfo{volume}{16}, \bibinfo{number}{1} (\bibinfo{year}{2019}),
  \bibinfo{pages}{175--194}.
\newblock
\newblock
\shownote{Publisher: Wiley Online Library.}


\bibitem[\protect\citeauthoryear{Berk, Heidari, Jabbari, Kearns, and Roth}{Berk
  et~al\mbox{.}}{2021}]%
        {berkFairnessCriminalJustice2021}
\bibfield{author}{\bibinfo{person}{Richard Berk}, \bibinfo{person}{Hoda
  Heidari}, \bibinfo{person}{Shahin Jabbari}, \bibinfo{person}{Michael Kearns},
  {and} \bibinfo{person}{Aaron Roth}.} \bibinfo{year}{2021}\natexlab{}.
\newblock \showarticletitle{Fairness in criminal justice risk assessments:
  {The} state of the art}.
\newblock \bibinfo{journal}{\emph{Sociological Methods \& Research}}
  \bibinfo{volume}{50}, \bibinfo{number}{1} (\bibinfo{year}{2021}),
  \bibinfo{pages}{3--44}.
\newblock
\newblock
\shownote{Publisher: Sage Publications Sage CA: Los Angeles, CA.}


\bibitem[\protect\citeauthoryear{Beutel, Chen, Zhao, and Chi}{Beutel
  et~al\mbox{.}}{2017}]%
        {beutelDataDecisionsTheoretical2017}
\bibfield{author}{\bibinfo{person}{Alex Beutel}, \bibinfo{person}{Jilin Chen},
  \bibinfo{person}{Zhe Zhao}, {and} \bibinfo{person}{Ed~H. Chi}.}
  \bibinfo{year}{2017}\natexlab{}.
\newblock \showarticletitle{Data {Decisions} and {Theoretical} {Implications}
  when {Adversarially} {Learning} {Fair} {Representations}}.
\newblock \bibinfo{journal}{\emph{arXiv:1707.00075 [cs]}} (\bibinfo{date}{July}
  \bibinfo{year}{2017}).
\newblock
\urldef\tempurl%
\url{http://arxiv.org/abs/1707.00075}
\showURL{%
\tempurl}
\newblock
\shownote{arXiv: 1707.00075.}


\bibitem[\protect\citeauthoryear{Bojarski, Del~Testa, Dworakowski, Firner,
  Flepp, Goyal, Jackel, Monfort, Muller, and Zhang}{Bojarski
  et~al\mbox{.}}{2016}]%
        {bojarskiEndEndLearning2016}
\bibfield{author}{\bibinfo{person}{Mariusz Bojarski}, \bibinfo{person}{Davide
  Del~Testa}, \bibinfo{person}{Daniel Dworakowski}, \bibinfo{person}{Bernhard
  Firner}, \bibinfo{person}{Beat Flepp}, \bibinfo{person}{Prasoon Goyal},
  \bibinfo{person}{Lawrence~D. Jackel}, \bibinfo{person}{Mathew Monfort},
  \bibinfo{person}{Urs Muller}, {and} \bibinfo{person}{Jiakai Zhang}.}
  \bibinfo{year}{2016}\natexlab{}.
\newblock \showarticletitle{End to end learning for self-driving cars}.
\newblock \bibinfo{journal}{\emph{arXiv preprint arXiv:1604.07316}}
  (\bibinfo{year}{2016}).
\newblock


\bibitem[\protect\citeauthoryear{Brennan and Oliver}{Brennan and
  Oliver}{2013}]%
        {brennanEmergenceMachineLearning2013}
\bibfield{author}{\bibinfo{person}{Tim Brennan} {and}
  \bibinfo{person}{William~L. Oliver}.} \bibinfo{year}{2013}\natexlab{}.
\newblock \showarticletitle{Emergence of machine learning techniques in
  criminology: implications of complexity in our data and in research
  questions}.
\newblock \bibinfo{journal}{\emph{Criminology \& Pub. Pol'y}}
  \bibinfo{volume}{12} (\bibinfo{year}{2013}), \bibinfo{pages}{551}.
\newblock
\newblock
\shownote{Publisher: HeinOnline.}


\bibitem[\protect\citeauthoryear{Caton and Haas}{Caton and Haas}{2020}]%
        {catonFairnessMachineLearning2020}
\bibfield{author}{\bibinfo{person}{Simon Caton} {and}
  \bibinfo{person}{Christian Haas}.} \bibinfo{year}{2020}\natexlab{}.
\newblock \showarticletitle{Fairness in {Machine} {Learning}: {A} {Survey}}.
\newblock \bibinfo{journal}{\emph{arXiv:2010.04053 [cs, stat]}}
  (\bibinfo{date}{Oct.} \bibinfo{year}{2020}).
\newblock
\urldef\tempurl%
\url{http://arxiv.org/abs/2010.04053}
\showURL{%
\tempurl}
\newblock
\shownote{arXiv: 2010.04053.}


\bibitem[\protect\citeauthoryear{Che, Li, Jacob, Bengio, and Li}{Che
  et~al\mbox{.}}{2017}]%
        {cheModeRegularizedGenerative2017}
\bibfield{author}{\bibinfo{person}{Tong Che}, \bibinfo{person}{Yanran Li},
  \bibinfo{person}{Athul~Paul Jacob}, \bibinfo{person}{Yoshua Bengio}, {and}
  \bibinfo{person}{Wenjie Li}.} \bibinfo{year}{2017}\natexlab{}.
\newblock \showarticletitle{Mode {Regularized} {Generative} {Adversarial}
  {Networks}}.
\newblock \bibinfo{journal}{\emph{arXiv:1612.02136 [cs]}}
  (\bibinfo{date}{March} \bibinfo{year}{2017}).
\newblock
\urldef\tempurl%
\url{http://arxiv.org/abs/1612.02136}
\showURL{%
\tempurl}
\newblock
\shownote{arXiv: 1612.02136.}


\bibitem[\protect\citeauthoryear{Chouldechova}{Chouldechova}{2017}]%
        {chouldechovaFairPredictionDisparate2017}
\bibfield{author}{\bibinfo{person}{Alexandra Chouldechova}.}
  \bibinfo{year}{2017}\natexlab{}.
\newblock \showarticletitle{Fair prediction with disparate impact: {A} study of
  bias in recidivism prediction instruments}.
\newblock \bibinfo{journal}{\emph{Big data}} \bibinfo{volume}{5},
  \bibinfo{number}{2} (\bibinfo{year}{2017}), \bibinfo{pages}{153--163}.
\newblock
\newblock
\shownote{Publisher: Mary Ann Liebert, Inc. 140 Huguenot Street, 3rd Floor New
  Rochelle, NY 10801 USA.}


\bibitem[\protect\citeauthoryear{Cleary}{Cleary}{1966}]%
        {clearyTestBiasValidity1966}
\bibfield{author}{\bibinfo{person}{T.~Anne Cleary}.}
  \bibinfo{year}{1966}\natexlab{}.
\newblock \showarticletitle{Test bias: {Validity} of the {Scholastic}
  {Aptitude} {Test} for {Negro} and {White} students in integrated colleges}.
\newblock \bibinfo{journal}{\emph{ETS Research Bulletin Series}}
  \bibinfo{volume}{1966}, \bibinfo{number}{2} (\bibinfo{year}{1966}),
  \bibinfo{pages}{i--23}.
\newblock
\newblock
\shownote{Publisher: Wiley Online Library.}


\bibitem[\protect\citeauthoryear{Delobelle, Temple, Perrouin, Frénay, Heymans,
  and Berendt}{Delobelle et~al\mbox{.}}{2021}]%
        {delobelleEthicalAdversariesMitigating2021}
\bibfield{author}{\bibinfo{person}{Pieter Delobelle}, \bibinfo{person}{Paul
  Temple}, \bibinfo{person}{Gilles Perrouin}, \bibinfo{person}{Benoît
  Frénay}, \bibinfo{person}{Patrick Heymans}, {and} \bibinfo{person}{Bettina
  Berendt}.} \bibinfo{year}{2021}\natexlab{}.
\newblock \showarticletitle{Ethical adversaries: {Towards} mitigating
  unfairness with adversarial machine learning}.
\newblock \bibinfo{journal}{\emph{ACM SIGKDD Explorations Newsletter}}
  \bibinfo{volume}{23}, \bibinfo{number}{1} (\bibinfo{year}{2021}),
  \bibinfo{pages}{32--41}.
\newblock
\newblock
\shownote{Publisher: ACM New York, NY, USA.}


\bibitem[\protect\citeauthoryear{Dwork, Hardt, Pitassi, Reingold, and
  Zemel}{Dwork et~al\mbox{.}}{2012}]%
        {dworkFairnessAwareness2012}
\bibfield{author}{\bibinfo{person}{Cynthia Dwork}, \bibinfo{person}{Moritz
  Hardt}, \bibinfo{person}{Toniann Pitassi}, \bibinfo{person}{Omer Reingold},
  {and} \bibinfo{person}{Richard Zemel}.} \bibinfo{year}{2012}\natexlab{}.
\newblock \showarticletitle{Fairness through awareness}. In
  \bibinfo{booktitle}{\emph{Proceedings of the 3rd {Innovations} in
  {Theoretical} {Computer} {Science} {Conference} on - {ITCS} '12}}.
  \bibinfo{publisher}{ACM Press}, \bibinfo{address}{Cambridge, Massachusetts},
  \bibinfo{pages}{214--226}.
\newblock
\showISBNx{978-1-4503-1115-1}
\urldef\tempurl%
\url{https://doi.org/10.1145/2090236.2090255}
\showDOI{\tempurl}


\bibitem[\protect\citeauthoryear{Elazar and Goldberg}{Elazar and
  Goldberg}{2018}]%
        {elazarAdversarialRemovalDemographic2018}
\bibfield{author}{\bibinfo{person}{Yanai Elazar} {and} \bibinfo{person}{Yoav
  Goldberg}.} \bibinfo{year}{2018}\natexlab{}.
\newblock \showarticletitle{Adversarial removal of demographic attributes from
  text data}.
\newblock \bibinfo{journal}{\emph{arXiv preprint arXiv:1808.06640}}
  (\bibinfo{year}{2018}).
\newblock


\bibitem[\protect\citeauthoryear{Feldman, Friedler, Moeller, Scheidegger, and
  Venkatasubramanian}{Feldman et~al\mbox{.}}{2015}]%
        {feldmanCertifyingRemovingDisparate2015}
\bibfield{author}{\bibinfo{person}{Michael Feldman},
  \bibinfo{person}{Sorelle~A. Friedler}, \bibinfo{person}{John Moeller},
  \bibinfo{person}{Carlos Scheidegger}, {and} \bibinfo{person}{Suresh
  Venkatasubramanian}.} \bibinfo{year}{2015}\natexlab{}.
\newblock \showarticletitle{Certifying and {Removing} {Disparate} {Impact}}. In
  \bibinfo{booktitle}{\emph{Proceedings of the 21th {ACM} {SIGKDD}
  {International} {Conference} on {Knowledge} {Discovery} and {Data} {Mining} -
  {KDD} '15}}. \bibinfo{publisher}{ACM Press}, \bibinfo{address}{Sydney, NSW,
  Australia}, \bibinfo{pages}{259--268}.
\newblock
\showISBNx{978-1-4503-3664-2}
\urldef\tempurl%
\url{https://doi.org/10.1145/2783258.2783311}
\showDOI{\tempurl}


\bibitem[\protect\citeauthoryear{Freeman, Shah, and Vaish}{Freeman
  et~al\mbox{.}}{2020}]%
        {freemanBestBothWorlds2020}
\bibfield{author}{\bibinfo{person}{Rupert Freeman}, \bibinfo{person}{Nisarg
  Shah}, {and} \bibinfo{person}{Rohit Vaish}.} \bibinfo{year}{2020}\natexlab{}.
\newblock \showarticletitle{Best of {Both} {Worlds}: {Ex}-{Ante} and
  {Ex}-{Post} {Fairness} in {Resource} {Allocation}}.
\newblock \bibinfo{journal}{\emph{arXiv:2005.14122 [cs]}} (\bibinfo{date}{May}
  \bibinfo{year}{2020}).
\newblock
\urldef\tempurl%
\url{http://arxiv.org/abs/2005.14122}
\showURL{%
\tempurl}
\newblock
\shownote{arXiv: 2005.14122.}


\bibitem[\protect\citeauthoryear{Friedler, Scheidegger, Venkatasubramanian,
  Choudhary, Hamilton, and Roth}{Friedler et~al\mbox{.}}{2019}]%
        {friedlerComparativeStudyFairnessenhancing2019a}
\bibfield{author}{\bibinfo{person}{Sorelle~A. Friedler},
  \bibinfo{person}{Carlos Scheidegger}, \bibinfo{person}{Suresh
  Venkatasubramanian}, \bibinfo{person}{Sonam Choudhary},
  \bibinfo{person}{Evan~P. Hamilton}, {and} \bibinfo{person}{Derek Roth}.}
  \bibinfo{year}{2019}\natexlab{}.
\newblock \showarticletitle{A comparative study of fairness-enhancing
  interventions in machine learning}. In \bibinfo{booktitle}{\emph{Proceedings
  of the {Conference} on {Fairness}, {Accountability}, and {Transparency}}}
  \emph{(\bibinfo{series}{{FAT}* '19})}. \bibinfo{publisher}{Association for
  Computing Machinery}, \bibinfo{address}{New York, NY, USA},
  \bibinfo{pages}{329--338}.
\newblock
\showISBNx{978-1-4503-6125-5}
\urldef\tempurl%
\url{https://doi.org/10.1145/3287560.3287589}
\showDOI{\tempurl}


\bibitem[\protect\citeauthoryear{Ganin and Lempitsky}{Ganin and
  Lempitsky}{[n.d.]}]%
        {ganinUnsupervisedDomainAdaptation}
\bibfield{author}{\bibinfo{person}{Yaroslav Ganin} {and}
  \bibinfo{person}{Victor Lempitsky}.} \bibinfo{year}{[n.d.]}\natexlab{}.
\newblock \showarticletitle{Unsupervised {Domain} {Adaptation} by
  {Backpropagation}}.
\newblock  (\bibinfo{year}{[n.\,d.]}), \bibinfo{pages}{10}.
\newblock


\bibitem[\protect\citeauthoryear{Gao}{Gao}{2022}]%
        {gaoFairNeuron2022}
\bibfield{author}{\bibinfo{person}{Xuanqi Gao}.}
  \bibinfo{year}{2022}\natexlab{}.
\newblock \bibinfo{title}{{FairNeuron}}.
\newblock
\newblock
\urldef\tempurl%
\url{https://github.com/Antimony5292/FairNeuron}
\showURL{%
\tempurl}
\newblock
\shownote{original-date: 2021-09-01T12:52:43Z.}


\bibitem[\protect\citeauthoryear{Goodfellow, Pouget-Abadie, Mirza, Xu,
  Warde-Farley, Ozair, Courville, and Bengio}{Goodfellow et~al\mbox{.}}{2014}]%
        {goodfellowGenerativeAdversarialNets2014}
\bibfield{author}{\bibinfo{person}{Ian Goodfellow}, \bibinfo{person}{Jean
  Pouget-Abadie}, \bibinfo{person}{Mehdi Mirza}, \bibinfo{person}{Bing Xu},
  \bibinfo{person}{David Warde-Farley}, \bibinfo{person}{Sherjil Ozair},
  \bibinfo{person}{Aaron Courville}, {and} \bibinfo{person}{Yoshua Bengio}.}
  \bibinfo{year}{2014}\natexlab{}.
\newblock \showarticletitle{Generative {Adversarial} {Nets}}. In
  \bibinfo{booktitle}{\emph{Advances in {Neural} {Information} {Processing}
  {Systems}}}, Vol.~\bibinfo{volume}{27}. \bibinfo{publisher}{Curran
  Associates, Inc.}
\newblock
\urldef\tempurl%
\url{https://papers.nips.cc/paper/2014/hash/5ca3e9b122f61f8f06494c97b1afccf3-Abstract.html}
\showURL{%
\tempurl}


\bibitem[\protect\citeauthoryear{Guion}{Guion}{1966}]%
        {guionEmploymentTestsDiscriminatory1966}
\bibfield{author}{\bibinfo{person}{Robert~M. Guion}.}
  \bibinfo{year}{1966}\natexlab{}.
\newblock \showarticletitle{Employment tests and discriminatory hiring}.
\newblock \bibinfo{journal}{\emph{Industrial Relations: A Journal of Economy
  and Society}} \bibinfo{volume}{5}, \bibinfo{number}{2}
  (\bibinfo{year}{1966}), \bibinfo{pages}{20--37}.
\newblock
\newblock
\shownote{Publisher: Wiley Online Library.}


\bibitem[\protect\citeauthoryear{Hardt, Price, and Srebro}{Hardt
  et~al\mbox{.}}{2016}]%
        {hardtEqualityOpportunitySupervised2016}
\bibfield{author}{\bibinfo{person}{Moritz Hardt}, \bibinfo{person}{Eric Price},
  {and} \bibinfo{person}{Nati Srebro}.} \bibinfo{year}{2016}\natexlab{}.
\newblock \showarticletitle{Equality of opportunity in supervised learning}.
\newblock \bibinfo{journal}{\emph{Advances in neural information processing
  systems}}  \bibinfo{volume}{29} (\bibinfo{year}{2016}),
  \bibinfo{pages}{3315--3323}.
\newblock


\bibitem[\protect\citeauthoryear{Hashimoto, Srivastava, Namkoong, and
  Liang}{Hashimoto et~al\mbox{.}}{2018}]%
        {hashimotoFairnessDemographicsRepeated2018}
\bibfield{author}{\bibinfo{person}{Tatsunori Hashimoto}, \bibinfo{person}{Megha
  Srivastava}, \bibinfo{person}{Hongseok Namkoong}, {and}
  \bibinfo{person}{Percy Liang}.} \bibinfo{year}{2018}\natexlab{}.
\newblock \showarticletitle{Fairness without demographics in repeated loss
  minimization}. In \bibinfo{booktitle}{\emph{International {Conference} on
  {Machine} {Learning}}}. \bibinfo{publisher}{PMLR},
  \bibinfo{pages}{1929--1938}.
\newblock


\bibitem[\protect\citeauthoryear{He, Zhang, Ren, and Sun}{He
  et~al\mbox{.}}{2016}]%
        {heDeepResidualLearning2016}
\bibfield{author}{\bibinfo{person}{Kaiming He}, \bibinfo{person}{Xiangyu
  Zhang}, \bibinfo{person}{Shaoqing Ren}, {and} \bibinfo{person}{Jian Sun}.}
  \bibinfo{year}{2016}\natexlab{}.
\newblock \showarticletitle{Deep residual learning for image recognition}. In
  \bibinfo{booktitle}{\emph{Proceedings of the {IEEE} conference on computer
  vision and pattern recognition}}. \bibinfo{pages}{770--778}.
\newblock


\bibitem[\protect\citeauthoryear{Hu, Lu, Li, and Chen}{Hu
  et~al\mbox{.}}{2014}]%
        {huConvolutionalNeuralNetwork2014}
\bibfield{author}{\bibinfo{person}{Baotian Hu}, \bibinfo{person}{Zhengdong Lu},
  \bibinfo{person}{Hang Li}, {and} \bibinfo{person}{Qingcai Chen}.}
  \bibinfo{year}{2014}\natexlab{}.
\newblock \showarticletitle{Convolutional neural network architectures for
  matching natural language sentences}. In \bibinfo{booktitle}{\emph{Advances
  in neural information processing systems}}. \bibinfo{pages}{2042--2050}.
\newblock


\bibitem[\protect\citeauthoryear{Kalchbrenner, Grefenstette, and
  Blunsom}{Kalchbrenner et~al\mbox{.}}{2014}]%
        {kalchbrennerConvolutionalNeuralNetwork2014}
\bibfield{author}{\bibinfo{person}{Nal Kalchbrenner}, \bibinfo{person}{Edward
  Grefenstette}, {and} \bibinfo{person}{Phil Blunsom}.}
  \bibinfo{year}{2014}\natexlab{}.
\newblock \showarticletitle{A convolutional neural network for modelling
  sentences}.
\newblock \bibinfo{journal}{\emph{arXiv preprint arXiv:1404.2188}}
  (\bibinfo{year}{2014}).
\newblock


\bibitem[\protect\citeauthoryear{Kamiran and Calders}{Kamiran and
  Calders}{2009}]%
        {kamiranClassifyingDiscriminating2009}
\bibfield{author}{\bibinfo{person}{Faisal Kamiran} {and} \bibinfo{person}{Toon
  Calders}.} \bibinfo{year}{2009}\natexlab{}.
\newblock \showarticletitle{Classifying without discriminating}. In
  \bibinfo{booktitle}{\emph{2009 2nd international conference on computer,
  control and communication}}. \bibinfo{publisher}{IEEE},
  \bibinfo{pages}{1--6}.
\newblock


\bibitem[\protect\citeauthoryear{Kamiran and Calders}{Kamiran and
  Calders}{2012}]%
        {kamiranDataPreprocessingTechniques2012}
\bibfield{author}{\bibinfo{person}{Faisal Kamiran} {and} \bibinfo{person}{Toon
  Calders}.} \bibinfo{year}{2012}\natexlab{}.
\newblock \showarticletitle{Data preprocessing techniques for classification
  without discrimination}.
\newblock \bibinfo{journal}{\emph{Knowledge and Information Systems}}
  \bibinfo{volume}{33}, \bibinfo{number}{1} (\bibinfo{date}{Oct.}
  \bibinfo{year}{2012}), \bibinfo{pages}{1--33}.
\newblock
\showISSN{0219-1377, 0219-3116}
\urldef\tempurl%
\url{https://doi.org/10.1007/s10115-011-0463-8}
\showDOI{\tempurl}


\bibitem[\protect\citeauthoryear{Kamiran, Karim, and Zhang}{Kamiran
  et~al\mbox{.}}{2012}]%
        {kamiranDecisionTheoryDiscriminationAware2012}
\bibfield{author}{\bibinfo{person}{Faisal Kamiran}, \bibinfo{person}{Asim
  Karim}, {and} \bibinfo{person}{Xiangliang Zhang}.}
  \bibinfo{year}{2012}\natexlab{}.
\newblock \showarticletitle{Decision {Theory} for {Discrimination}-{Aware}
  {Classification}}. In \bibinfo{booktitle}{\emph{2012 {IEEE} 12th
  {International} {Conference} on {Data} {Mining}}}. \bibinfo{publisher}{IEEE},
  \bibinfo{address}{Brussels, Belgium}, \bibinfo{pages}{924--929}.
\newblock
\showISBNx{978-1-4673-4649-8 978-0-7695-4905-7}
\urldef\tempurl%
\url{https://doi.org/10.1109/ICDM.2012.45}
\showDOI{\tempurl}


\bibitem[\protect\citeauthoryear{Kleinberg, Ludwig, Mullainathan, and
  Rambachan}{Kleinberg et~al\mbox{.}}{2018}]%
        {kleinbergAlgorithmicFairness2018}
\bibfield{author}{\bibinfo{person}{Jon Kleinberg}, \bibinfo{person}{Jens
  Ludwig}, \bibinfo{person}{Sendhil Mullainathan}, {and}
  \bibinfo{person}{Ashesh Rambachan}.} \bibinfo{year}{2018}\natexlab{}.
\newblock \showarticletitle{Algorithmic fairness}. In
  \bibinfo{booktitle}{\emph{Aea papers and proceedings}},
  Vol.~\bibinfo{volume}{108}. \bibinfo{pages}{22--27}.
\newblock


\bibitem[\protect\citeauthoryear{Kusner, Loftus, Russell, and Silva}{Kusner
  et~al\mbox{.}}{[n.d.]}]%
        {kusnerCounterfactualFairness}
\bibfield{author}{\bibinfo{person}{Matt~J Kusner}, \bibinfo{person}{Joshua
  Loftus}, \bibinfo{person}{Chris Russell}, {and} \bibinfo{person}{Ricardo
  Silva}.} \bibinfo{year}{[n.d.]}\natexlab{}.
\newblock \showarticletitle{Counterfactual {Fairness}}.
\newblock \bibinfo{journal}{\emph{NIPS 2017}} (\bibinfo{year}{[n.\,d.]}),
  \bibinfo{pages}{11}.
\newblock


\bibitem[\protect\citeauthoryear{Lahoti, Beutel, Chen, Lee, Prost, Thain, Wang,
  and Chi}{Lahoti et~al\mbox{.}}{2020}]%
        {lahotiFairnessDemographicsAdversarially2020}
\bibfield{author}{\bibinfo{person}{Preethi Lahoti}, \bibinfo{person}{Alex
  Beutel}, \bibinfo{person}{Jilin Chen}, \bibinfo{person}{Kang Lee},
  \bibinfo{person}{Flavien Prost}, \bibinfo{person}{Nithum Thain},
  \bibinfo{person}{Xuezhi Wang}, {and} \bibinfo{person}{Ed~H. Chi}.}
  \bibinfo{year}{2020}\natexlab{}.
\newblock \showarticletitle{Fairness without {Demographics} through
  {Adversarially} {Reweighted} {Learning}}.
\newblock \bibinfo{journal}{\emph{arXiv:2006.13114 [cs, stat]}}
  (\bibinfo{date}{Nov.} \bibinfo{year}{2020}).
\newblock
\urldef\tempurl%
\url{http://arxiv.org/abs/2006.13114}
\showURL{%
\tempurl}
\newblock
\shownote{arXiv: 2006.13114.}


\bibitem[\protect\citeauthoryear{Lahoti, Gummadi, and Weikum}{Lahoti
  et~al\mbox{.}}{2019}]%
        {lahotiOperationalizingIndividualFairness2019}
\bibfield{author}{\bibinfo{person}{Preethi Lahoti}, \bibinfo{person}{Krishna~P.
  Gummadi}, {and} \bibinfo{person}{Gerhard Weikum}.}
  \bibinfo{year}{2019}\natexlab{}.
\newblock \showarticletitle{Operationalizing individual fairness with pairwise
  fair representations}.
\newblock \bibinfo{journal}{\emph{arXiv preprint arXiv:1907.01439}}
  (\bibinfo{year}{2019}).
\newblock


\bibitem[\protect\citeauthoryear{Ma, Aafer, Xu, Lee, Zhai, Liu, and Zhang}{Ma
  et~al\mbox{.}}{2017}]%
        {maLAMPDataProvenance2017}
\bibfield{author}{\bibinfo{person}{Shiqing Ma}, \bibinfo{person}{Yousra Aafer},
  \bibinfo{person}{Zhaogui Xu}, \bibinfo{person}{Wen-Chuan Lee},
  \bibinfo{person}{Juan Zhai}, \bibinfo{person}{Yingqi Liu}, {and}
  \bibinfo{person}{Xiangyu Zhang}.} \bibinfo{year}{2017}\natexlab{}.
\newblock \showarticletitle{{LAMP}: data provenance for graph based machine
  learning algorithms through derivative computation}. In
  \bibinfo{booktitle}{\emph{Proceedings of the 2017 11th {Joint} {Meeting} on
  {Foundations} of {Software} {Engineering}}}. \bibinfo{pages}{786--797}.
\newblock


\bibitem[\protect\citeauthoryear{Ma, Liu, Lee, Zhang, and Grama}{Ma
  et~al\mbox{.}}{2018}]%
        {maMODEAutomatedNeural2018}
\bibfield{author}{\bibinfo{person}{Shiqing Ma}, \bibinfo{person}{Yingqi Liu},
  \bibinfo{person}{Wen-Chuan Lee}, \bibinfo{person}{Xiangyu Zhang}, {and}
  \bibinfo{person}{Ananth Grama}.} \bibinfo{year}{2018}\natexlab{}.
\newblock \showarticletitle{{MODE}: automated neural network model debugging
  via state differential analysis and input selection}. In
  \bibinfo{booktitle}{\emph{Proceedings of the 2018 26th {ACM} {Joint}
  {Meeting} on {European} {Software} {Engineering} {Conference} and {Symposium}
  on the {Foundations} of {Software} {Engineering} - {ESEC}/{FSE} 2018}}.
  \bibinfo{publisher}{ACM Press}, \bibinfo{address}{Lake Buena Vista, FL, USA},
  \bibinfo{pages}{175--186}.
\newblock
\showISBNx{978-1-4503-5573-5}
\urldef\tempurl%
\url{https://doi.org/10.1145/3236024.3236082}
\showDOI{\tempurl}


\bibitem[\protect\citeauthoryear{Markham and Buchanan}{Markham and
  Buchanan}{2012}]%
        {markhamEthicalDecisionmakingInternet2012}
\bibfield{author}{\bibinfo{person}{Annette Markham} {and}
  \bibinfo{person}{Elizabeth Buchanan}.} \bibinfo{year}{2012}\natexlab{}.
\newblock \showarticletitle{Ethical decision-making and internet research:
  {Version} 2.0. recommendations from the {AoIR} ethics working committee}.
\newblock \bibinfo{journal}{\emph{Available online: aoir. org/reports/ethics2.
  pdf}} (\bibinfo{year}{2012}).
\newblock


\bibitem[\protect\citeauthoryear{Mertikopoulos, Papadimitriou, and
  Piliouras}{Mertikopoulos et~al\mbox{.}}{2018}]%
        {mertikopoulosCyclesAdversarialRegularized2018}
\bibfield{author}{\bibinfo{person}{Panayotis Mertikopoulos},
  \bibinfo{person}{Christos Papadimitriou}, {and} \bibinfo{person}{Georgios
  Piliouras}.} \bibinfo{year}{2018}\natexlab{}.
\newblock \showarticletitle{Cycles in {Adversarial} {Regularized} {Learning}}.
\newblock In \bibinfo{booktitle}{\emph{Proceedings of the 2018 {Annual}
  {ACM}-{SIAM} {Symposium} on {Discrete} {Algorithms} ({SODA})}}.
  \bibinfo{publisher}{Society for Industrial and Applied Mathematics},
  \bibinfo{pages}{2703--2717}.
\newblock
\urldef\tempurl%
\url{https://doi.org/10.1137/1.9781611975031.172}
\showDOI{\tempurl}


\bibitem[\protect\citeauthoryear{O'neil}{O'neil}{2016}]%
        {oneilWeaponsMathDestruction2016}
\bibfield{author}{\bibinfo{person}{Cathy O'neil}.}
  \bibinfo{year}{2016}\natexlab{}.
\newblock \bibinfo{booktitle}{\emph{Weapons of math destruction: {How} big data
  increases inequality and threatens democracy}}.
\newblock \bibinfo{publisher}{Crown}.
\newblock


\bibitem[\protect\citeauthoryear{Paszke, Gross, Massa, Lerer, Bradbury, Chanan,
  Killeen, Lin, Gimelshein, Antiga, Desmaison, Kopf, Yang, DeVito, Raison,
  Tejani, Chilamkurthy, Steiner, Fang, Bai, and Chintala}{Paszke
  et~al\mbox{.}}{[n.d.]}]%
        {paszkePyTorchImperativeStyle}
\bibfield{author}{\bibinfo{person}{Adam Paszke}, \bibinfo{person}{Sam Gross},
  \bibinfo{person}{Francisco Massa}, \bibinfo{person}{Adam Lerer},
  \bibinfo{person}{James Bradbury}, \bibinfo{person}{Gregory Chanan},
  \bibinfo{person}{Trevor Killeen}, \bibinfo{person}{Zeming Lin},
  \bibinfo{person}{Natalia Gimelshein}, \bibinfo{person}{Luca Antiga},
  \bibinfo{person}{Alban Desmaison}, \bibinfo{person}{Andreas Kopf},
  \bibinfo{person}{Edward Yang}, \bibinfo{person}{Zachary DeVito},
  \bibinfo{person}{Martin Raison}, \bibinfo{person}{Alykhan Tejani},
  \bibinfo{person}{Sasank Chilamkurthy}, \bibinfo{person}{Benoit Steiner},
  \bibinfo{person}{Lu Fang}, \bibinfo{person}{Junjie Bai}, {and}
  \bibinfo{person}{Soumith Chintala}.} \bibinfo{year}{[n.d.]}\natexlab{}.
\newblock \showarticletitle{{PyTorch}: {An} {Imperative} {Style},
  {High}-{Performance} {Deep} {Learning} {Library}}.
\newblock  (\bibinfo{year}{[n.\,d.]}), \bibinfo{pages}{12}.
\newblock


\bibitem[\protect\citeauthoryear{Pleiss, Raghavan, Wu, Kleinberg, and
  Weinberger}{Pleiss et~al\mbox{.}}{[n.d.]}]%
        {pleissFairnessCalibration}
\bibfield{author}{\bibinfo{person}{Geoff Pleiss}, \bibinfo{person}{Manish
  Raghavan}, \bibinfo{person}{Felix Wu}, \bibinfo{person}{Jon Kleinberg}, {and}
  \bibinfo{person}{Kilian~Q Weinberger}.} \bibinfo{year}{[n.d.]}\natexlab{}.
\newblock \showarticletitle{On {Fairness} and {Calibration}}.
\newblock  (\bibinfo{year}{[n.\,d.]}), \bibinfo{pages}{10}.
\newblock


\bibitem[\protect\citeauthoryear{President, Munoz, Director, Science, Policy)),
  Policy, Science, and Policy))}{President et~al\mbox{.}}{2016}]%
        {presidentBigDataReport2016}
\bibfield{author}{\bibinfo{person}{Executive Office of~the President},
  \bibinfo{person}{Cecilia Munoz}, \bibinfo{person}{Domestic Policy~Council
  Director}, \bibinfo{person}{Megan (US Chief Technology Officer Smith
  (Office~of Science}, \bibinfo{person}{Technology Policy))},
  \bibinfo{person}{DJ~(Deputy Chief Technology Officer for~Data Policy},
  \bibinfo{person}{Chief Data Scientist Patil (Office~of Science}, {and}
  \bibinfo{person}{Technology Policy))}.} \bibinfo{year}{2016}\natexlab{}.
\newblock \bibinfo{booktitle}{\emph{Big data: {A} report on algorithmic
  systems, opportunity, and civil rights}}.
\newblock \bibinfo{publisher}{Executive Office of the President}.
\newblock


\bibitem[\protect\citeauthoryear{President and Podesta}{President and
  Podesta}{2014}]%
        {presidentBigDataSeizing2014}
\bibfield{author}{\bibinfo{person}{United States Executive Office of~the
  President} {and} \bibinfo{person}{John Podesta}.}
  \bibinfo{year}{2014}\natexlab{}.
\newblock \bibinfo{booktitle}{\emph{Big data: {Seizing} opportunities,
  preserving values}}.
\newblock \bibinfo{publisher}{White House, Executive Office of the President}.
\newblock


\bibitem[\protect\citeauthoryear{Qiu, Leng, Guo, Chen, Li, Guo, and Zhu}{Qiu
  et~al\mbox{.}}{2019}]%
        {qiuAdversarialDefenseNetwork2019}
\bibfield{author}{\bibinfo{person}{Yuxian Qiu}, \bibinfo{person}{Jingwen Leng},
  \bibinfo{person}{Cong Guo}, \bibinfo{person}{Quan Chen},
  \bibinfo{person}{Chao Li}, \bibinfo{person}{Minyi Guo}, {and}
  \bibinfo{person}{Yuhao Zhu}.} \bibinfo{year}{2019}\natexlab{}.
\newblock \showarticletitle{Adversarial {Defense} {Through} {Network}
  {Profiling} {Based} {Path} {Extraction}}. In \bibinfo{booktitle}{\emph{2019
  {IEEE}/{CVF} {Conference} on {Computer} {Vision} and {Pattern} {Recognition}
  ({CVPR})}}. \bibinfo{publisher}{IEEE}, \bibinfo{address}{Long Beach, CA,
  USA}, \bibinfo{pages}{4772--4781}.
\newblock
\showISBNx{978-1-72813-293-8}
\urldef\tempurl%
\url{https://doi.org/10.1109/CVPR.2019.00491}
\showDOI{\tempurl}


\bibitem[\protect\citeauthoryear{Raghavan, Barocas, Kleinberg, and
  Levy}{Raghavan et~al\mbox{.}}{2020}]%
        {raghavanMitigatingBiasAlgorithmic2020}
\bibfield{author}{\bibinfo{person}{Manish Raghavan}, \bibinfo{person}{Solon
  Barocas}, \bibinfo{person}{Jon Kleinberg}, {and} \bibinfo{person}{Karen
  Levy}.} \bibinfo{year}{2020}\natexlab{}.
\newblock \showarticletitle{Mitigating bias in algorithmic hiring: {Evaluating}
  claims and practices}. In \bibinfo{booktitle}{\emph{Proceedings of the 2020
  conference on fairness, accountability, and transparency}}.
  \bibinfo{pages}{469--481}.
\newblock


\bibitem[\protect\citeauthoryear{Sattigeri, Hoffman, Chenthamarakshan, and
  Varshney}{Sattigeri et~al\mbox{.}}{2019}]%
        {sattigeriFairnessGANGenerating2019}
\bibfield{author}{\bibinfo{person}{Prasanna Sattigeri},
  \bibinfo{person}{Samuel~C. Hoffman}, \bibinfo{person}{Vijil
  Chenthamarakshan}, {and} \bibinfo{person}{Kush~R. Varshney}.}
  \bibinfo{year}{2019}\natexlab{}.
\newblock \showarticletitle{Fairness {GAN}: {Generating} datasets with fairness
  properties using a generative adversarial network}.
\newblock \bibinfo{journal}{\emph{IBM Journal of Research and Development}}
  \bibinfo{volume}{63}, \bibinfo{number}{4/5} (\bibinfo{year}{2019}),
  \bibinfo{pages}{3--1}.
\newblock
\newblock
\shownote{Publisher: IBM.}


\bibitem[\protect\citeauthoryear{Tramer, Atlidakis, Geambasu, Hsu, Hubaux,
  Humbert, Juels, and Lin}{Tramer et~al\mbox{.}}{2017}]%
        {tramerFairTestDiscoveringUnwarranted2017}
\bibfield{author}{\bibinfo{person}{Florian Tramer}, \bibinfo{person}{Vaggelis
  Atlidakis}, \bibinfo{person}{Roxana Geambasu}, \bibinfo{person}{Daniel Hsu},
  \bibinfo{person}{Jean-Pierre Hubaux}, \bibinfo{person}{Mathias Humbert},
  \bibinfo{person}{Ari Juels}, {and} \bibinfo{person}{Huang Lin}.}
  \bibinfo{year}{2017}\natexlab{}.
\newblock \showarticletitle{{FairTest}: {Discovering} {Unwarranted}
  {Associations} in {Data}-{Driven} {Applications}}. In
  \bibinfo{booktitle}{\emph{2017 {IEEE} {European} {Symposium} on {Security}
  and {Privacy} ({EuroS}\&{P})}}. \bibinfo{publisher}{IEEE},
  \bibinfo{address}{Paris}, \bibinfo{pages}{401--416}.
\newblock
\showISBNx{978-1-5090-5762-7}
\urldef\tempurl%
\url{https://doi.org/10.1109/EuroSP.2017.29}
\showDOI{\tempurl}


\bibitem[\protect\citeauthoryear{Udeshi, Arora, and Chattopadhyay}{Udeshi
  et~al\mbox{.}}{2018}]%
        {udeshiAutomatedDirectedFairness2018}
\bibfield{author}{\bibinfo{person}{Sakshi Udeshi}, \bibinfo{person}{Pryanshu
  Arora}, {and} \bibinfo{person}{Sudipta Chattopadhyay}.}
  \bibinfo{year}{2018}\natexlab{}.
\newblock \showarticletitle{Automated directed fairness testing}. In
  \bibinfo{booktitle}{\emph{Proceedings of the 33rd {ACM}/{IEEE}
  {International} {Conference} on {Automated} {Software} {Engineering} - {ASE}
  2018}}. \bibinfo{publisher}{ACM Press}, \bibinfo{address}{Montpellier,
  France}, \bibinfo{pages}{98--108}.
\newblock
\showISBNx{978-1-4503-5937-5}
\urldef\tempurl%
\url{https://doi.org/10.1145/3238147.3238165}
\showDOI{\tempurl}


\bibitem[\protect\citeauthoryear{van~den Broek, Sergeeva, and Huysman}{van~den
  Broek et~al\mbox{.}}{2019}]%
        {vandenbroekHiringAlgorithmsEthnography2019}
\bibfield{author}{\bibinfo{person}{Elmira van~den Broek},
  \bibinfo{person}{Anastasia Sergeeva}, {and} \bibinfo{person}{Marleen
  Huysman}.} \bibinfo{year}{2019}\natexlab{}.
\newblock \showarticletitle{Hiring algorithms: {An} ethnography of fairness in
  practice}.
\newblock  (\bibinfo{year}{2019}).
\newblock


\bibitem[\protect\citeauthoryear{Voigt and Von~dem Bussche}{Voigt and Von~dem
  Bussche}{2017}]%
        {voigtEuGeneralData2017}
\bibfield{author}{\bibinfo{person}{Paul Voigt} {and} \bibinfo{person}{Axel
  Von~dem Bussche}.} \bibinfo{year}{2017}\natexlab{}.
\newblock \showarticletitle{The eu general data protection regulation (gdpr)}.
\newblock \bibinfo{journal}{\emph{A Practical Guide, 1st Ed., Cham: Springer
  International Publishing}}  \bibinfo{volume}{10} (\bibinfo{year}{2017}),
  \bibinfo{pages}{3152676}.
\newblock
\newblock
\shownote{Publisher: Springer.}


\bibitem[\protect\citeauthoryear{Wang, Su, Zhang, and Hu}{Wang
  et~al\mbox{.}}{2018}]%
        {wangInterpretNeuralNetworks2018}
\bibfield{author}{\bibinfo{person}{Yulong Wang}, \bibinfo{person}{Hang Su},
  \bibinfo{person}{Bo Zhang}, {and} \bibinfo{person}{Xiaolin Hu}.}
  \bibinfo{year}{2018}\natexlab{}.
\newblock \showarticletitle{Interpret {Neural} {Networks} by {Identifying}
  {Critical} {Data} {Routing} {Paths}}. In \bibinfo{booktitle}{\emph{2018
  {IEEE}/{CVF} {Conference} on {Computer} {Vision} and {Pattern}
  {Recognition}}}. \bibinfo{publisher}{IEEE}, \bibinfo{address}{Salt Lake City,
  UT}, \bibinfo{pages}{8906--8914}.
\newblock
\showISBNx{978-1-5386-6420-9}
\urldef\tempurl%
\url{https://doi.org/10.1109/CVPR.2018.00928}
\showDOI{\tempurl}


\bibitem[\protect\citeauthoryear{Xu, Yuan, Zhang, and Wu}{Xu
  et~al\mbox{.}}{2018}]%
        {xuFairganFairnessawareGenerative2018}
\bibfield{author}{\bibinfo{person}{Depeng Xu}, \bibinfo{person}{Shuhan Yuan},
  \bibinfo{person}{Lu Zhang}, {and} \bibinfo{person}{Xintao Wu}.}
  \bibinfo{year}{2018}\natexlab{}.
\newblock \showarticletitle{Fairgan: {Fairness}-aware generative adversarial
  networks}. In \bibinfo{booktitle}{\emph{2018 {IEEE} {International}
  {Conference} on {Big} {Data} ({Big} {Data})}}. \bibinfo{publisher}{IEEE},
  \bibinfo{pages}{570--575}.
\newblock


\bibitem[\protect\citeauthoryear{Zafar, Valera, Gomez~Rodriguez, and
  Gummadi}{Zafar et~al\mbox{.}}{2017a}]%
        {zafarFairnessDisparateTreatment2017}
\bibfield{author}{\bibinfo{person}{Muhammad~Bilal Zafar},
  \bibinfo{person}{Isabel Valera}, \bibinfo{person}{Manuel Gomez~Rodriguez},
  {and} \bibinfo{person}{Krishna~P. Gummadi}.}
  \bibinfo{year}{2017}\natexlab{a}.
\newblock \showarticletitle{Fairness {Beyond} {Disparate} {Treatment} \&amp;
  {Disparate} {Impact}: {Learning} {Classification} without {Disparate}
  {Mistreatment}}. In \bibinfo{booktitle}{\emph{Proceedings of the 26th
  {International} {Conference} on {World} {Wide} {Web}}}
  \emph{(\bibinfo{series}{{WWW} '17})}. \bibinfo{publisher}{International World
  Wide Web Conferences Steering Committee}, \bibinfo{address}{Republic and
  Canton of Geneva, CHE}, \bibinfo{pages}{1171--1180}.
\newblock
\showISBNx{978-1-4503-4913-0}
\urldef\tempurl%
\url{https://doi.org/10.1145/3038912.3052660}
\showDOI{\tempurl}


\bibitem[\protect\citeauthoryear{Zafar, Valera, Rogriguez, and Gummadi}{Zafar
  et~al\mbox{.}}{2017b}]%
        {zafarFairnessConstraintsMechanisms2017}
\bibfield{author}{\bibinfo{person}{Muhammad~Bilal Zafar},
  \bibinfo{person}{Isabel Valera}, \bibinfo{person}{Manuel~Gomez Rogriguez},
  {and} \bibinfo{person}{Krishna~P. Gummadi}.}
  \bibinfo{year}{2017}\natexlab{b}.
\newblock \showarticletitle{Fairness constraints: {Mechanisms} for fair
  classification}. In \bibinfo{booktitle}{\emph{Artificial {Intelligence} and
  {Statistics}}}. \bibinfo{publisher}{PMLR}, \bibinfo{pages}{962--970}.
\newblock


\bibitem[\protect\citeauthoryear{Zemel}{Zemel}{[n.d.]}]%
        {zemelLearningFairRepresentations}
\bibfield{author}{\bibinfo{person}{Richard Zemel}.}
  \bibinfo{year}{[n.d.]}\natexlab{}.
\newblock \showarticletitle{Learning {Fair} {Representations}}.
\newblock  (\bibinfo{year}{[n.\,d.]}), \bibinfo{pages}{9}.
\newblock


\bibitem[\protect\citeauthoryear{Zhang, Lemoine, and Mitchell}{Zhang
  et~al\mbox{.}}{2018}]%
        {zhangMitigatingUnwantedBiases2018}
\bibfield{author}{\bibinfo{person}{Brian~Hu Zhang}, \bibinfo{person}{Blake
  Lemoine}, {and} \bibinfo{person}{Margaret Mitchell}.}
  \bibinfo{year}{2018}\natexlab{}.
\newblock \showarticletitle{Mitigating {Unwanted} {Biases} with {Adversarial}
  {Learning}}. In \bibinfo{booktitle}{\emph{Proceedings of the 2018
  {AAAI}/{ACM} {Conference} on {AI}, {Ethics}, and {Society}}}.
  \bibinfo{publisher}{ACM}, \bibinfo{address}{New Orleans LA USA},
  \bibinfo{pages}{335--340}.
\newblock
\showISBNx{978-1-4503-6012-8}
\urldef\tempurl%
\url{https://doi.org/10.1145/3278721.3278779}
\showDOI{\tempurl}


\bibitem[\protect\citeauthoryear{Zhang and Shah}{Zhang and Shah}{2014}]%
        {zhangFairnessMultiagentSequential2014}
\bibfield{author}{\bibinfo{person}{Chongjie Zhang} {and}
  \bibinfo{person}{Julie~A. Shah}.} \bibinfo{year}{2014}\natexlab{}.
\newblock \showarticletitle{Fairness in multi-agent sequential
  decision-making}. In \bibinfo{booktitle}{\emph{Advances in {Neural}
  {Information} {Processing} {Systems}}}. \bibinfo{pages}{2636--2644}.
\newblock


\bibitem[\protect\citeauthoryear{Zhang, Wu, and Wu}{Zhang
  et~al\mbox{.}}{2017}]%
        {zhangAchievingNonDiscriminationData2017}
\bibfield{author}{\bibinfo{person}{Lu Zhang}, \bibinfo{person}{Yongkai Wu},
  {and} \bibinfo{person}{Xintao Wu}.} \bibinfo{year}{2017}\natexlab{}.
\newblock \showarticletitle{Achieving {Non}-{Discrimination} in {Data}
  {Release}}. In \bibinfo{booktitle}{\emph{Proceedings of the 23rd {ACM}
  {SIGKDD} {International} {Conference} on {Knowledge} {Discovery} and {Data}
  {Mining}}}. \bibinfo{publisher}{ACM}, \bibinfo{address}{Halifax NS Canada},
  \bibinfo{pages}{1335--1344}.
\newblock
\showISBNx{978-1-4503-4887-4}
\urldef\tempurl%
\url{https://doi.org/10.1145/3097983.3098167}
\showDOI{\tempurl}


\bibitem[\protect\citeauthoryear{Zhang, Wang, Sun, Dong, Wang, Wang, Dong, and
  Dai}{Zhang et~al\mbox{.}}{2020b}]%
        {zhangWhiteboxFairnessTesting2020}
\bibfield{author}{\bibinfo{person}{Peixin Zhang}, \bibinfo{person}{Jingyi
  Wang}, \bibinfo{person}{Jun Sun}, \bibinfo{person}{Guoliang Dong},
  \bibinfo{person}{Xinyu Wang}, \bibinfo{person}{Xingen Wang},
  \bibinfo{person}{Jin~Song Dong}, {and} \bibinfo{person}{Ting Dai}.}
  \bibinfo{year}{2020}\natexlab{b}.
\newblock \showarticletitle{White-box fairness testing through adversarial
  sampling}. In \bibinfo{booktitle}{\emph{Proceedings of the {ACM}/{IEEE} 42nd
  {International} {Conference} on {Software} {Engineering}}}.
  \bibinfo{publisher}{ACM}, \bibinfo{address}{Seoul South Korea},
  \bibinfo{pages}{949--960}.
\newblock
\showISBNx{978-1-4503-7121-6}
\urldef\tempurl%
\url{https://doi.org/10.1145/3377811.3380331}
\showDOI{\tempurl}


\bibitem[\protect\citeauthoryear{Zhang, Zhai, Ma, and Shen}{Zhang
  et~al\mbox{.}}{2021}]%
        {zhangAUTOTRAINERAutomaticDNN2021}
\bibfield{author}{\bibinfo{person}{Xiaoyu Zhang}, \bibinfo{person}{Juan Zhai},
  \bibinfo{person}{Shiqing Ma}, {and} \bibinfo{person}{Chao Shen}.}
  \bibinfo{year}{2021}\natexlab{}.
\newblock \showarticletitle{{AUTOTRAINER}: {An} {Automatic} {DNN} {Training}
  {Problem} {Detection} and {Repair} {System}}. In
  \bibinfo{booktitle}{\emph{2021 {IEEE}/{ACM} 43rd {International} {Conference}
  on {Software} {Engineering} ({ICSE})}}. \bibinfo{publisher}{IEEE},
  \bibinfo{address}{Madrid, ES}, \bibinfo{pages}{359--371}.
\newblock
\showISBNx{978-1-66540-296-5}
\urldef\tempurl%
\url{https://doi.org/10.1109/ICSE43902.2021.00043}
\showDOI{\tempurl}


\bibitem[\protect\citeauthoryear{Zhang, Li, Guo, Chen, and Liu}{Zhang
  et~al\mbox{.}}{2020a}]%
        {zhangDynamicSlicingDeep2020}
\bibfield{author}{\bibinfo{person}{Ziqi Zhang}, \bibinfo{person}{Yuanchun Li},
  \bibinfo{person}{Yao Guo}, \bibinfo{person}{Xiangqun Chen}, {and}
  \bibinfo{person}{Yunxin Liu}.} \bibinfo{year}{2020}\natexlab{a}.
\newblock \showarticletitle{Dynamic {Slicing} for {Deep} {Neural} {Networks}}.
\newblock  (\bibinfo{year}{2020}), \bibinfo{pages}{13}.
\newblock


\end{thebibliography}

\end{document}